\documentclass{article}
\usepackage{log_2023}						

\usepackage{booktabs}						
\usepackage{multirow}						
\usepackage{amsfonts}						
\usepackage{graphicx}						
\usepackage{duckuments}						

\usepackage[numbers,compress,sort]{natbib}	

\title[A Generalization Benchmark for GNNs Predicting Fluid Dynamics]{SURF: A Generalization Benchmark for GNNs Predicting Fluid Dynamics}

\author[S. Künzli et al.]{%
Stefan Künzli\\
ETH Zurich\\
\email{skuenzli@ethz.ch}
\And
Florian Grötschla\thanks{Equal contribution.}\\
ETH Zurich\\
\email{fgroetschla@ethz.ch}
\And
Joël Mathys\footnotemark[1]\\
ETH Zurich\\
\email{jmathys@ethz.ch}
\And
Roger Wattenhofer\\
ETH Zurich\\
\email{wattenhofer@ethz.ch}
}

\begin{document}

\maketitle
\begin{abstract}
    Simulating fluid dynamics is crucial for the design and development process, ranging from simple valves to complex turbomachinery.
    Accurately solving the underlying physical equations is computationally expensive. Therefore, learning-based solvers that model interactions on meshes have gained interest due to their promising speed-ups. However, it is unknown to what extent these models truly understand the underlying physical principles and can generalize rather than interpolate. Generalization is a key requirement for a general-purpose fluid simulator, which should adapt to different topologies, resolutions, or thermodynamic ranges. We propose SURF, a benchmark designed to test the \textit{generalization} of learned graph-based fluid simulators. SURF comprises individual datasets and provides specific performance and generalization metrics for evaluating and comparing different models. We empirically demonstrate the applicability of SURF by thoroughly investigating the two state-of-the-art graph-based models, yielding new insights into their generalization. SURF is available under \url{https://github.com/s-kuenzli/surf-fluidsimulation}. 
\end{abstract}

\section{Introduction}

Fluid simulations have established themselves as indispensable tools for addressing intricate and complex tasks in design and development processes. They play a central role in the construction of Formula 1 cars, where fluid simulations are extensively harnessed to refine external aerodynamics and enhance the performance of combustion engines~\cite{formula1}. Moreover, they enable the accurate simulation of scenarios that are too expensive or even infeasible to conduct in real-world experiments, such as the examination of atmospheric re-entry vehicles~\cite{reentry}. The application of fluid simulations extends even to domains such as weather prediction~\cite{zajaczkowski2011preliminary}.

The underlying physics governing these phenomena can be modeled by the Navier-Stokes equations. Unfortunately, no closed-form solution is known, and they remain one of the infamous ``Millennium Problems''~\cite{millennium}. Thus, computationally expensive iterative numerical solutions of the non-linear partial differential equations are used.
The ongoing development of simulation software is geared towards delivering high-quality outcomes in the shortest feasible time frame~\cite{nvidia}.
Here, learning-based approaches have the potential to accelerate this development by either directly predicting the simulated process or providing a better initial solution that can be refined by classical iterative solvers~\cite{wiewel2019latent}. Often, this comes with a direct tradeoff between guaranteed precision and rapid processing time. 

Due to the common modeling of the simulation environment as irregular meshes, Graph Neural Network (GNN) based approaches have gained substantial traction recently. 
There, the main research focus lies on \textit{how accurately GNNs can predict} the flow of fluid dynamics. The aim is to create a learned general-purpose fluid simulator. 
Given the wide range of applications of such a simulator, one of its most essential abilities is to generalize and accurately predict the fluid dynamics in novel and unseen environments as they might arise in the aforementioned development processes. This poses a significant challenge, as training on all possible environments or scenarios is simply infeasible. As a consequence, such a model has to truly grasp the underlying physical principles in order to generalize rather than only interpolate between observed training data points. To be more precise, such a simulator should be able to adapt to different topologies, resolutions, and thermodynamic ranges. Therefore, we want to address the question \textit{how well GNNs can generalize} to different fluid dynamics environments. 

We propose SURF, a benchmark for the simulation of fluid dynamics that focuses explicitly on generalization capabilities to evaluate how well a model grasps the underlying physical principles. 
SURF contains a diverse set of training and benchmarking datasets, with the explicit aim of adapting to (1) finer mesh resolution, (2) different parameter ranges, (3) topologies, and (4) dynamic simulation environments.
To quantify the generalization, we propose a new set of metrics, the SURF generalization scores, measuring the performance of each mentioned aspect. They intend to capture the relative performance loss of a model due to not training on an unseen scenario where it has to rely on its ability to generalize.
To demonstrate the utility and applicability of SURF, we empirically evaluate two state-of-the-art models, MGN \cite{MGNPaper} and EAGLE \cite{EaglePaper}, for mesh-based fluid simulation. 

We propose the SURF benchmark to systematically evaluate the ability of learned mesh-based simulators to generalize across the demanding environments of fluid dynamics. SURF's main contributions can be summarized as follows:

\begin{itemize}
    \item We introduce seven new large-scale datasets to simulate the fluid dynamics behavior. Each dataset is constructed to change specific aspects of the environment to assess generalization performance across different datasets. 
    \item We propose the SURF generalization scores. A set of new metrics which quantitatively capture the ability of a proposed model architecture to generalize to overall or specific changes across mesh resolution, parameter ranges, topologies, or simulation dynamics. This is complemented by the SURF performance score, which can assess the prediction quality of a model.
    \item We investigate what aspects of the training process help models to generalize by identifying how a training data set should be built such that the generalization of a model improves.
    \item We conduct a thorough empirical evaluation and comparison of two state-of-the-art GNN-based fluid simulators concerning their ability to generalize.
\end{itemize}

\begin{figure}
	\centering
	\includegraphics[width=0.9\linewidth]{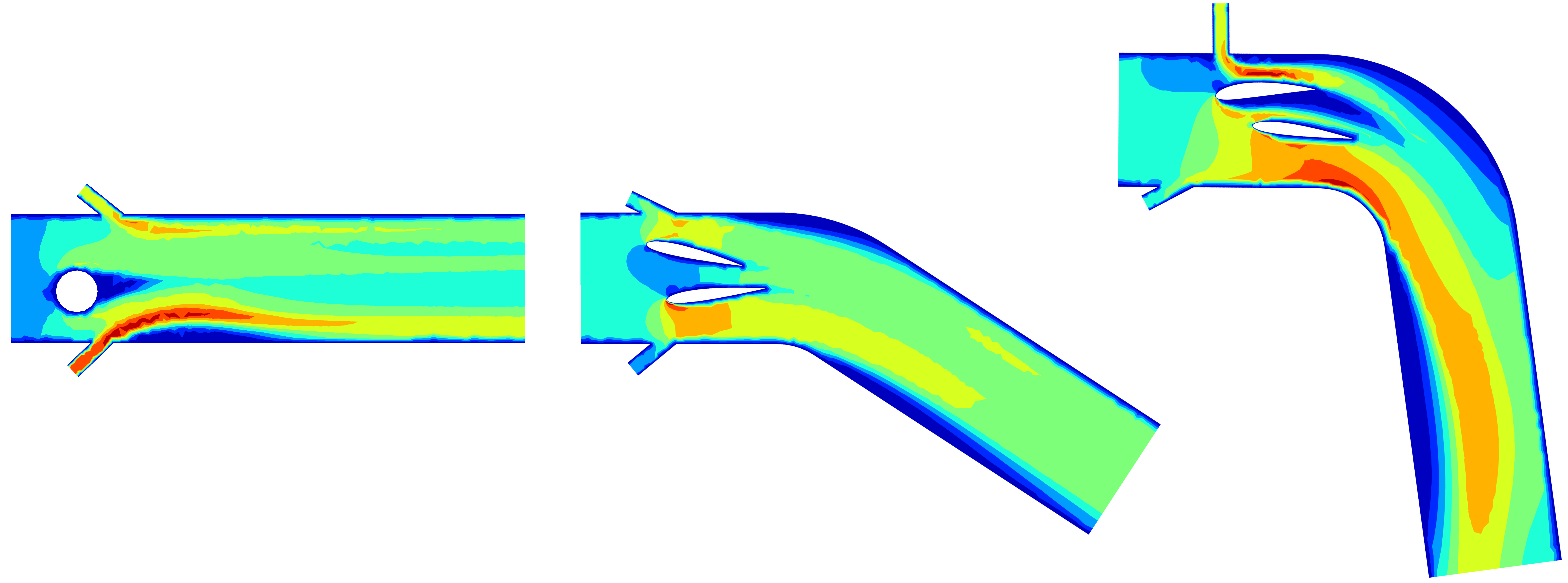}
	\caption{Velocity visualization for three datapoints of the SURF benchmark from SURF-Base, SURF-Topology, and SURF-Full dataset, respectively. The datasets are designed to measure a specific aspect of generalization, e.g., how well models can adapt to new objects, such as different air foils or unseen topologies with turns. Values are displayed contoured to make visual comparisons easier.}
	\label{fig:storyline}
\end{figure}

\section{Related Work}

In recent years there has been much interest in deploying machine learning for fluid simulation. The different approaches can be divided into the following categories: particle-based, grid-based, mesh-based and operator learning methods. The particle and grid methods are described in the Appendix~\ref{sec:ExtendedRelatedWork}, whereas the mesh-based and operator learning methods are described in the following.

\paragraph{Mesh-Based Methods}

In industrial contexts, mesh-based fluid dynamics solvers are vital for simulating complex fluid behavior. Due to their ability to handle complex geometries with irregular meshes. Renowned solvers like ANSYS Fluent and Star-CCM+ or the open-source solver OpenFoam are extensively employed. In aerospace engineering, they are used to optimize aircraft designs for enhanced efficiency and safety. The automotive industry relies on them to simulate airflow around vehicles, reducing drag and improving fuel efficiency. These mesh-based solvers underpin industrial innovation by providing insights into fluid dynamics phenomena crucial for optimal design and performance across diverse sectors.

Meshes directly translate to graph representations, making them amenable to Graph Neural Networks (GNNs). In this context, mesh nodes and edges correspond to graph nodes and undirected edges within the graph structure.
Graph Element Networks~\cite{alet2019graph} use this structure, while works that combine graph structures with PDEs also exist~\cite{belbute2020combining}.
The most relevant architectures for our evaluation are MeshGraphNets (MGN)~\cite{MGNPaper} and EAGLE~\cite{EaglePaper}. 

MGN~\cite{MGNPaper} follows an encode-process-decode architecture and undergoes testing across diverse physical domains, including scenarios such as 2D incompressible fluid flow around a cylinder or cloth simulation. An extensive series of 1200 distinct simulations are conducted to compile a comprehensive dataset. Systematically varied parameters like inlet velocity, cylinder position, and radius are explored within each simulation, spanning 600 discrete time steps.

EAGLE~\cite{EaglePaper} improves upon this approach by introducing enhancements such as node clustering, graph pooling, and global attention mechanisms. These refinements strengthen the model's ability to capture global dependencies. Simultaneously, a  dataset featuring 1,184 simulations depicting 2D unsteady fluid flow originating from a mobile flow source across 600 distinct fluid domains was generated.

\paragraph{Operator Learning}
A different approach for predicting solutions of partial differential equations like the Navier Stokes equations are machine learning methods which learn an operator mapping from the input functions to the solutions~\cite{LuDeepONet2019}. Several different neural architectures are investigated for learning operators. Fourier neural operator (FNO)~\cite{ZongyiLi2020} learns the operator in spectral space or transformer models~\cite{li2023transformer} which are based on attention mechanism. Nevertheless, the application of these methods to real-world problems is challenging. For example methods like FNO using Fast Fourier Transform require a uniform regular grid. These challenges are addressed in recent publications. Geo-FNO~\cite{li2022fourier} extends FNO by learning a mapping from an irregular mesh to a uniform mesh. Nevertheless, the adaption to general topologies remains a challenge. A General Neural Operator Transformer (GNOT)~\cite{HaoGNOT2023} for Operator Learning uses heterogeneous normalized attention layer, which is able to handle multiple input functions and irregular meshes.

\paragraph{Fluid Simulation Datasets}
Both MGN and EAGLE focus their evaluation on synthetic in-distribution datasets but also consider generalization as an essential quality that is discussed further. EAGLE investigates generalization to different ground topologies and mesh resolutions. MGN changes mesh sizes and tests on airfoils with steeper angles or with higher inflow speeds. 
More synthetic datasets for the purpose of learning were introduced~\cite{chen2021graph, han2022predicting}, for example for turbulence simulation~\cite{stachenfeld2021learned}. The PDEBench~\cite{takamoto2023pdebench} provides solutions to a wide variety of partial differntial equations (PDE), including a single dataset for the Navier-Stokes equation in a unit square with random initial velocities and forcing terms.

None of these works offer dedicated datasets or comprehensive scores designed to compare models for this specific purpose. Our evaluation bridges this gap, creating a unified benchmark to quantitatively analyze their respective generalization capabilities. In addition to velocity and pressure, we add temperature as a solution variable. This enables the models to calculate the fluid-solid heat transfer, an important process for optimizing heat exchangers or analyzing electronic cooling. This also allows the simulation of different thermal fluid properties resulting in different heat propagation in the fluid flow.

\section{Preliminaries}
\label{preliminaries}

\begin{figure}
	\centering
	\includegraphics[width=1.0\linewidth]{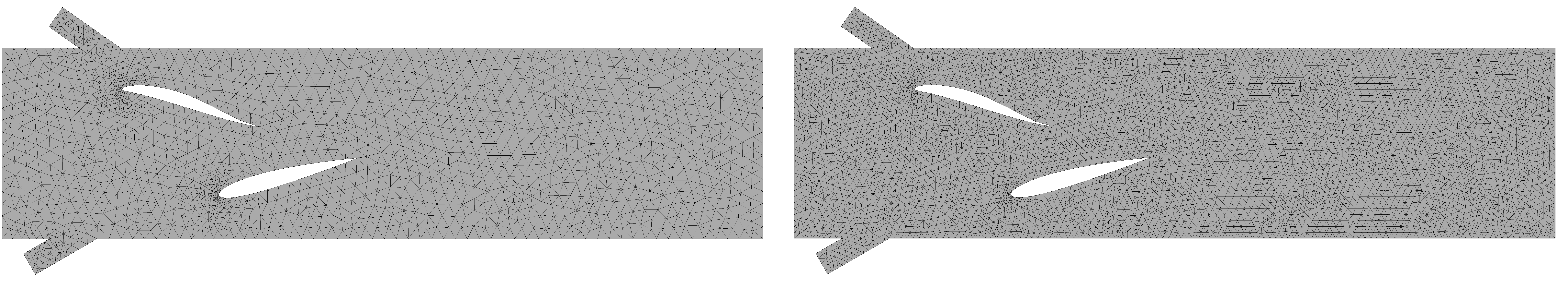}
	\caption{
 A mesh for a regular datapoint is visualized on the left, while the right datapoint depicts an instance from SURF-Mesh, with an increased mesh resolution by a factor of two.
 }
	\label{fig:mesh}
\end{figure}

\paragraph{Mesh Generation}
The geometry is built using Ansys SpaceClaim 2023 R1 \cite{ansysSpaceClaim2023R1}. Two different meshes are created with Ansys Meshing~\cite{ansysMeshing2023R1}. The fine mesh consists mostly of quadrilaterals and a small amount of triangles. At the wall boundaries, an inflation layer with eight elements is defined for a better resolution of the gradient. This mesh is then used for the ground truth calculation. The coarse mesh, consisting of triangles, is used for training and predictions (see Figure~\ref{fig:mesh}). The exact meshing algorithms used in Ansys Meshing are proprietary, but the triangular meshing of 2D geometries is a well-studied subject \cite{Bern1970}. The algorithms differ in the resulting mesh quality (the minimal angle of the triangle should be maximized), the number of created triangles for a certain geometry, and the generation time. Similar meshes could also be obtained with the open source mesher presented in~\cite{shewchuk96b}.
The result values (velocities, pressure, temperature) at the nodes of the coarse mesh are interpolated from the nodes of the fine mesh using linear interpolation on a triangular grid.

\paragraph{Root Mean Square Error (RMSE)}
To compare the predictions of a given model, the root mean square error (RMSE) is calculated for velocity, pressure, and temperature. The RMSE for the velocity $v$ for a time horizon $H$ on the test set $\mathcal{D}$ is calculated as:
\begin{equation*}
    \text{RMSE}_v(x,y) = \frac{1}{|\mathcal{D}|H} \sum_d^{\mathcal{D}}\sum_{t=1}^H \frac{1}{\sqrt{2}N_d} \sum_{v}^{N_d} ||\textbf{x}^t_{v} - \textbf{y}^t_{v}||_2 \label{eq:RMSE}
\end{equation*}
Where $\textbf{x}^t_{v}$ is the predicted velocity vector at time step $t$ at node $v$ and $\textbf{y}^t_{v}$ the corresponding ground truth value. $N_d$ denotes the number of nodes of the datapoint $d$. The RMSE for pressure $p$ and temperature $t$ are calculated analogously.

\section{SURF Benchmark}

We introduce SURF, a 2D fluid dynamics benchmark specifically tailored to assess generalization. The benchmark consists of seven new large-scale datasets accompanied by a custom set of metrics to quantify performance and generalization of proposed model architectures. In this section, we describe the design and generation process of the individual datasets and their intended purpose. Then, we define the SURF performance and generalization scores, which allow for a comprehensive evaluation and comparison against other baselines. Our dataset is publically available online\footnote{\url{https://huggingface.co/datasets/SURF-FluidSimulation/FluidSimulation}} with the accompanying code\footnote{\url{https://github.com/s-kuenzli/surf-fluidsimulation}} that includes evaluation scripts.

\begin{figure}
    \centering
    \includegraphics[width=1.0\linewidth]{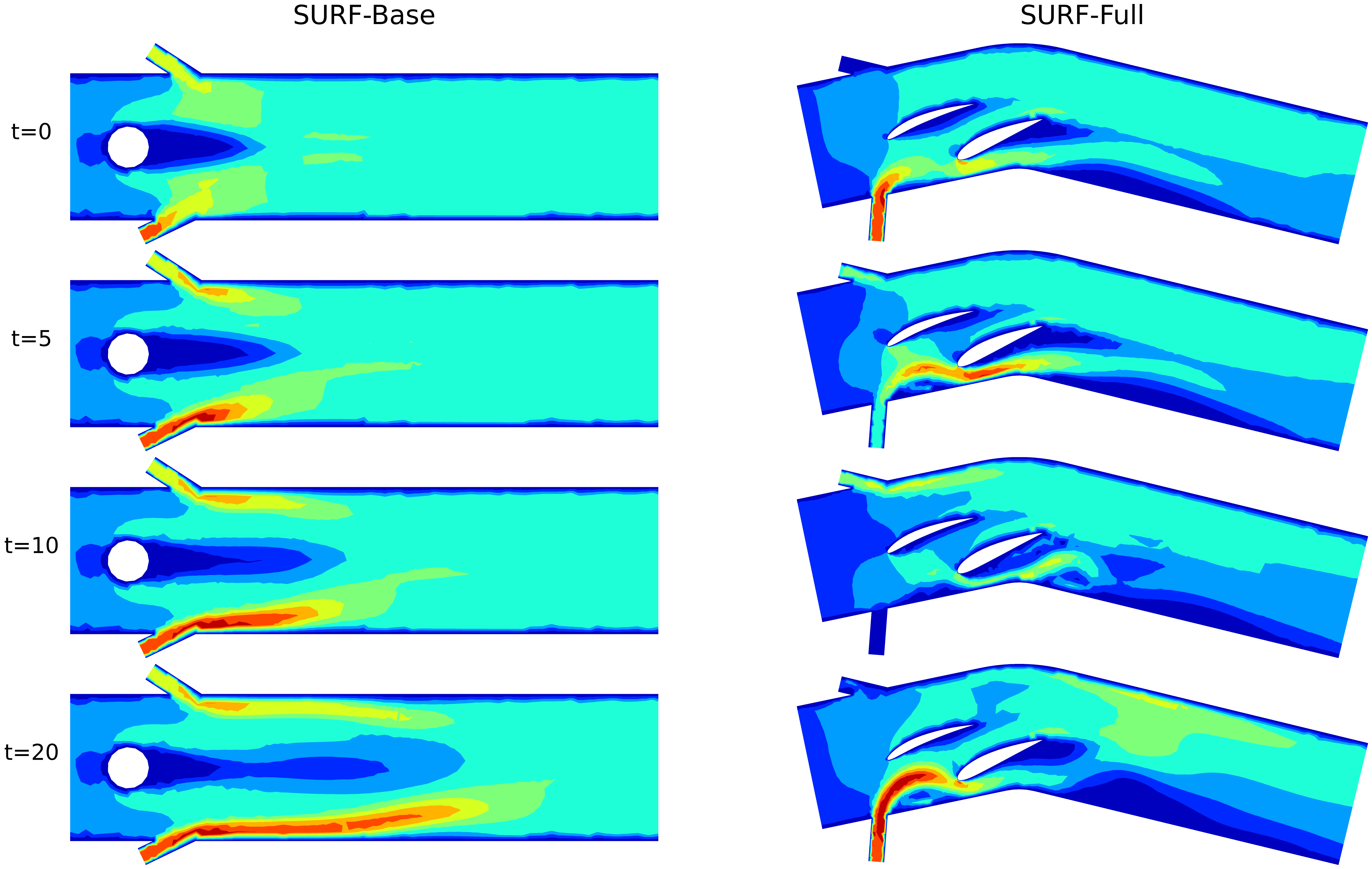}
    \caption{Velocity visualization for ground-truth data from SURF-Base (left) and SURF-Full (right), evaluated at timesteps 0, 5, 10, 20 (top to bottom). Note that on the right, the bottom inlet is active in the beginning, then turned off and turned on again in the last step, leading to a more dynamic simulation environment. Values are displayed contoured to make visual comparisons easier.}
    \label{fig:DatasetsOverview}
\end{figure}

\subsection{SURF Datasets} \label{chapter:Datasets}
Each of the seven SURF datasets is a large-scale collection of 2D fluid flow simulations with at least 1200 individual datapoints. They are accompanied by dedicated splits so that 80 percent are used for training and 10 percent for validation and test data each. The ground truth for each simulation is computed using ANSYS Fluent \cite{ansysHelpFluent}, a mesh-based solver using the finite volume method~\cite{eymard2000finite}. Afterward, the solution calculated on the fine mesh is downsampled using the linear interpolation described in Chapter \ref{preliminaries}. Every datapoint consists of $300$ timesteps and about $1800$ nodes that comprise the mesh topology. For the individual parameters, which determine the generation of each dataset, we refer to the Appendix. In the following, we outline the main differences between each of the individual datasets.

\paragraph{SURF-Base} This dataset serves as a basis for all other SURF datasets. It comprises simulated data of 2D flow around a single cylinder which can vary in size with two additional inlets, parametrized through position, angle, inflow velocity, and temperature. 
\paragraph{SURF-Rotated} The complete computational domain is rotated at an angle of [0°, 360°]. This ensures equal velocity distribution in all directions instead of the predominant velocity in the positive x-direction.
\paragraph{SURF-Range} Compared to the base dataset, the range of most simulation parameters is increased and decreased by a factor of two, including inlet velocity, object wall temperature, the radius of the cylinder, thermal conductivity, and heat capacity. In particular, this impacts the range of the calculated Reynolds number.
\paragraph{SURF-Topology} Two objects are present instead of a single cylinder. Each object is either a cylinder or one of five different airfoils, which vary in their angle of attack and can redirect the flow. The domain is equipped with an elbow angle from 0° to 90°. 
A flow separation may occur due to the velocity and the elbow angle.
\paragraph{SURF-Dynamic} Both small inlets vary their velocities over time. This introduces more variation of the calculated flow because, without the transient boundary conditions, the flow can converge to a steady state. Instead, the flow varies during the complete time horizon of the simulation.
\paragraph{SURF-Full} The superset of all other dataset generation parameters, except for datapoints with higher mesh resolutions. Therefore, it includes more temporal dynamics, larger ranges, multiple objects with diverse airfoils, and an elbow in the domain. Note that it contains datapoints not present in the other datasets as it is a superset of the generation parameters rather than the actual datapoints.
\paragraph{SURF-Mesh} Generated the same way as the full dataset, but the ground truth is downsampled to a finer mesh. The element size is reduced by a factor of two.

\subsection{SURF Metrics}
\label{sec:surf_metrics}
To measure the prediction quality of models on the SURF benchmark, we define a performance score $\text{PS}$. It follows the convention of the N-RMSE metric introduced in \cite{EaglePaper} and extends it to include temperature. The time horizon is always fixed at 250 timesteps.
\begin{equation*}
    \text{SURF}_\text{PS} = \frac{1}{3}(\text{PS}_v + \text{PS}_p + \text{PS}_t) = \frac{1}{3}\left( \frac{\text{RMSE}_v}{\sigma_v}+ \frac{\text{RMSE}_p}{\sigma_p} + \frac{\text{RMSE}_t}{\sigma_t}\right)
\end{equation*}
Where $\sigma_v$ denotes the standard deviation of the velocity on the respective dataset. Unless explicitly stated otherwise, the performance score is always reported on the $\text{SURF-Full}$ dataset.

The main aim of the SURF Benchmark is to quantify the ability of a model architecture to generalize across different environment settings. We measure generalization as the relative decrease in the performance of a model on the respective dataset. The idea is to train a model on an origin dataset and then test how well the performance translates to a different target dataset that differs in one aspect of the environment.
Formally, let $\text{GS}_v(\Phi, \mathcal{D}_{1}, \mathcal{D}_{2})$ denote the velocity generalization of model $\Phi$ across datasets $\mathcal{D}_1$ and $\mathcal{D}_2$. 
\begin{equation*}
    \text{GS}_{v}(\Phi, \mathcal{D}_1, \mathcal{D}_2) = \frac{\text{RMSE}_{v} \left( \Phi_{\mathcal{D}_1}(\mathcal{D}_2), \mathcal{D}_2 \right)} {\text{RMSE}_{v} \left( \Phi_{\mathcal{D}_2}(\mathcal{D}_2), \mathcal{D}_2 \right)}
\end{equation*}
Where $\Phi_{\mathcal{D}_1}(\mathcal{D}_2)$ denotes the prediction for the test set of $\mathcal{D}_2$ made by model $\Phi$ trained on the train set $\mathcal{D}_1$. The GS scores for temperature and pressure are calculated analogously. Moreover, the full generalization score is the average over all physical quantities: 
\begin{equation*}
    \text{GS} = \frac{1}{3}(\text{GS}_\text{v} + \text{GS}_\text{p} + \text{GS}_\text{t})
\end{equation*}

We identify four aspects fundamental to the ability to generalize: mesh resolution, parameter ranges, topologies, and dynamic simulation environments. Therefore, for each of these, we define a corresponding SURF generalization score, indicating how well the generalization of a single aspect of the environment has been learned:

\begin{align*}
    \text{SURF}_\text{GS-Mesh} (\Phi) &= \text{GS}(\Phi, \text{SURF-Full}, \text{SURF-Mesh})\\
    \text{SURF}_\text{GS-Topology} (\Phi) &= \text{GS}(\Phi, \text{SURF-Base}, \text{SURF-Topology})\\
    \text{SURF}_\text{GS-Range} (\Phi) &= \text{GS}(\Phi, \text{SURF-Base}, \text{SURF-Range})\\
    \text{SURF}_\text{GS-Dynamic} (\Phi) &= \text{GS}(\Phi, \text{SURF-Base}, \text{SURF-Dynamic})\\
\end{align*}
For assessing how well a model can generalize to different topologies, such as handling new objects, we can quantify this using the $\text{SURF}_\text{GS-Topology}$ score. To simplify evaluation and comparison across different models, we define the $\text{SURF}_\text{GS}$ score to capture the generalization capabilities of all aspects as an average over the more specialized generalization scores.  
\begin{equation*}
    \text{SURF}_\text{GS} = \frac{1}{4} \left(\text{SURF}_\text{GS-Mesh} + \text{SURF}_\text{GS-Topology} + \text{SURF}_\text{GS-Range} + \text{SURF}_\text{GS-Dynamic}\right)
\end{equation*}

\section{Evaluation}
We empirically evaluate two state-of-the-art models for fluid simulation, MeshGraphNets (MGN)~\cite{MGNPaper} and EAGLE~\cite{EaglePaper}. While MGN uses a message-passing Graph Neural Network to compute embeddings directly on the given mesh, EAGLE adds a global attention mechanism to enable long-range information exchange.
Both models undergo the SURF benchmark to assess their (1) baseline performance score, (2) generalization score, and (3) what training dataset generalizes best. 

We adjust both architectures slightly to incorporate the temperature as a parameter, which was not used in the original architectures and thus assumed constant. We use the same loss function for both MGN and EAGLE. More information on the exact training setup can be found in Appendix~\ref{sec:app_training}. All scores are reported as mean values over three runs. 

\subsection{Performance}
\begin{table}
    \centering
    \caption{Performance scores for velocity, pressure, and temperature after 250 time steps of MGN and EAGLE trained and tested on the SURF-Base, SURF-Rotated, and SURF-Full dataset, respectively. Both models achieve similar performance, but EAGLE predicts the pressure dynamics more accurately than MGN.}

    \begin{tabular}{cccccc}
        \toprule
        \textbf{Dataset} & \textbf{Model} & \textbf{PS$_\text{v}$ $\downarrow$} & \textbf{PS$_\text{p}$ $\downarrow$} & \textbf{PS$_\text{t}$ $\downarrow$} & \textbf{PS  $\downarrow$} \\
        \midrule
        Base & MGN & 0.073 $\pm$ 0.005 &  0.142 $\pm$ 0.011 &  0.068 $\pm$ 0.003 & 0.094 $\pm$ 0.006  \\ \cmidrule{2-6}
             & EAGLE & 0.074 $\pm$ 0.003 & 0.090 $\pm$ 0.004 &  0.097 $\pm$ 0.006 & 0.087 $\pm$ 0.004 \\ \cmidrule{1-6}
        Rotated & MGN & 0.041 $\pm$ 0.002 &  0.133 $\pm$ 0.003 & 0.071 $\pm$0.002 & 0.082 $\pm$ 0.001  \\ \cmidrule{2-6}
               & EAGLE & 0.061 $\pm$ 0.007 & 0.089 $\pm$ 0.005 &  0.107 $\pm$ 0.003 & 0.086 $\pm$ 0.004 \\ \cmidrule{1-6}
        Full & MGN & 0.165 $\pm$ 0.099  &  0.168 $\pm$ 0.088 &  0.111 $\pm$ 0.041 & 0.148 $\pm$ 0.076  \\ \cmidrule{2-6}
        & EAGLE & 0.163 $\pm$ 0.004&  0.166 $\pm$ 0.011 &  0.146 $\pm$ 0.015 & 0.158 $\pm$ 0.006  \\
        \bottomrule
    \end{tabular}
    \label{table:PS}
\end{table}

To assess the overall accuracy and quality of the predictions, we compare the results of MGN and EAGLE, across three distinct SURF datasets: Base, Rotated, and Full. The performance scores for each model and dataset are summarized in Table~\ref{table:PS}.

\paragraph{Velocity} Across all datasets, both MGN and EAGLE models demonstrate comparable performance in predicting velocity. Their scores consistently remain close, indicating a strong and balanced predictive capability in this aspect.

\paragraph{Pressure} EAGLE showcases a notable advantage in pressure prediction across the board. It consistently scores better than MGN, suggesting that EAGLE is better equipped to predict pressure-related outcomes accurately.

\paragraph{Temperature} For temperature prediction, the MGN model emerges as the more proficient one. Across all datasets, MGN consistently outperforms EAGLE in this aspect, suggesting a greater aptitude for capturing temperature-related patterns.

\paragraph{Overall Assessment} Considering all velocity, pressure, and temperature scores together, there is no clear winner between MGN and EAGLE. Both models achieve comparable performance, with EAGLE excelling in pressure prediction and MGN performing better in temperature prediction. This balanced performance across diverse aspects underscores the comparable overall performance of the two models.
Furthermore, the performance trends observed in the datasets remain consistent with the models' behavior. This consistency across datasets reinforces the stability of the introduced performance characteristics. Notably, the SURF-Full dataset consistently leads to the highest prediction loss, indicating that the increased complexity of this dataset might pose a challenge for both models. 
\begin{table}
    \centering
    \caption{Generalization scores after 250 time steps of the MGN and EAGLE model trained on the SURF-Base, SURF-Rotated, and SURF-Dynamic and then tested on the test set SURF-Topology, SURF-Range, SURF-Dynamic, and SURF-Full. Note that this is different from the SURF generalization scores.}
    \begin{tabular}{cccccc}
        \toprule
        \textbf{Training Set} & \textbf{Model} & \textbf{Topology $\downarrow$} & \textbf{Range $\downarrow$} & \textbf{Dynamic $\downarrow$} & \textbf{Full $\downarrow$}\\ \midrule
        Base & MGN & $ 3.10 \pm 0.72$ & 1.30 $\pm$ 0.07 & $ 5.76 \pm 0.64$ & $ 3.32 \pm 1.29$\\ \cmidrule{2-6}
        & EAGLE & $ 3.68 \pm 0.10$ & 1.08 $\pm$ 0.09 & $ 2.17 \pm 0.08$ & $ 3.73 \pm 0.52$ \\ \cmidrule{1-6}
        Rotated & MGN & 1.81 $\pm$ 0.38 & $ 1.54 \pm 0.18$ & $ 6.73 \pm 0.60$& 1.95 $\pm$ 0.72\\ \cmidrule{2-6}
        & EAGLE & 1.96 $\pm$ 0.08 & $ 1.36 \pm 0.22$ & $ 2.41 \pm 0.17$ & 1.47 $\pm$ 0.06 \\ \cmidrule{1-6}
        Dynamic & MGN & $ 3.11 \pm 0.56$ & $ 1.52 \pm 0.23$ & 1.00 $\pm$ 0.00& $ 3.93 \pm 1.52$ \\ \cmidrule{2-6}
        & EAGLE & $ 4.47 \pm 0.24$ & $ 2.16 \pm 0.33$ & 1.00 $\pm$ 0.00& $ 5.25 \pm 1.76$ \\
        \bottomrule
    \end{tabular}
    \label{table:GS_Combined}
\end{table}

\subsection{Generalization}
We investigate the ability of MGN and EAGLE models to generalize across individual aspects such as topologies, mesh resolution, parameter ranges, and simulation dynamics. We use the proposed SURF generalization scores defined in Section \ref{sec:surf_metrics}. Recall that the generalization score captures the relative drop in performance due to a model not training on the target dataset and therefore having to rely on its ability to generalize.
\begin{table}
\caption{The computed SURF generalization scores for MGN and EAGLE. While MGN can generalize better to different topologies, EAGLE achieves better generalization on mesh resolution, simulation dynamics, and parameter ranges, resulting in an overall better $\text{SURF}_{\text{GS}}$ score.}
    \centering
    \resizebox{\columnwidth}{!}{
    \begin{tabular}{cccccc}
    \toprule
        \textbf{Model} & \textbf{$\text{SURF}_\text{GS-Mesh}$ $\downarrow$} & \textbf{$\text{SURF}_\text{GS-Topology}$ $\downarrow$} & \textbf{$\text{SURF}_\text{GS-Range}$ $\downarrow$} & \textbf{$\text{SURF}_\text{Dynamic}$ $\downarrow$} & \textbf{$\text{SURF}_\text{GS}$ $\downarrow$}\\ \midrule
        MGN  & 1.07 $\pm$ 0.26 & 3.10 $\pm$ 0.72 & 1.30 $\pm$ 0.07 & 5.76 $\pm$ 0.64 & 2.81 $\pm$ 0.23 \\ \midrule
        EAGLE & 1.01 $\pm$ 0.05 & 3.68 $\pm$ 0.10 & 1.08 $\pm$ 0.09 & 2.17 $\pm$ 0.08 & 1.98 $\pm$ 0.06 \\ \bottomrule 
    \end{tabular}
    }
    \label{table:GS_Eagle_MGN}
\end{table}

The computed SURF generalization scores for each aspect are presented in Table \ref{table:GS_Eagle_MGN}. Our observation indicates that MGN exhibits better adaptability to unseen topologies compared to EAGLE. However, in an overall assessment, the EAGLE architecture demonstrates superior generalization regarding mesh resolution, parameter ranges, and simulation dynamics, which results in a lower overall SURF$_{\text{GS}}$ score.

\paragraph{Impact of Training Dataset on Generalization}

We explore what factors enhance training to yield better generalization results in the mentioned architectures. We analyze the GS scores, reflecting the efficacy of inter-dataset training-to-testing transfer. The results are shown in Table~\ref{table:GS_Combined}, and for the complete matrix, we refer to the appendix.

\begin{figure}
	\centering
	\includegraphics[width=\linewidth]{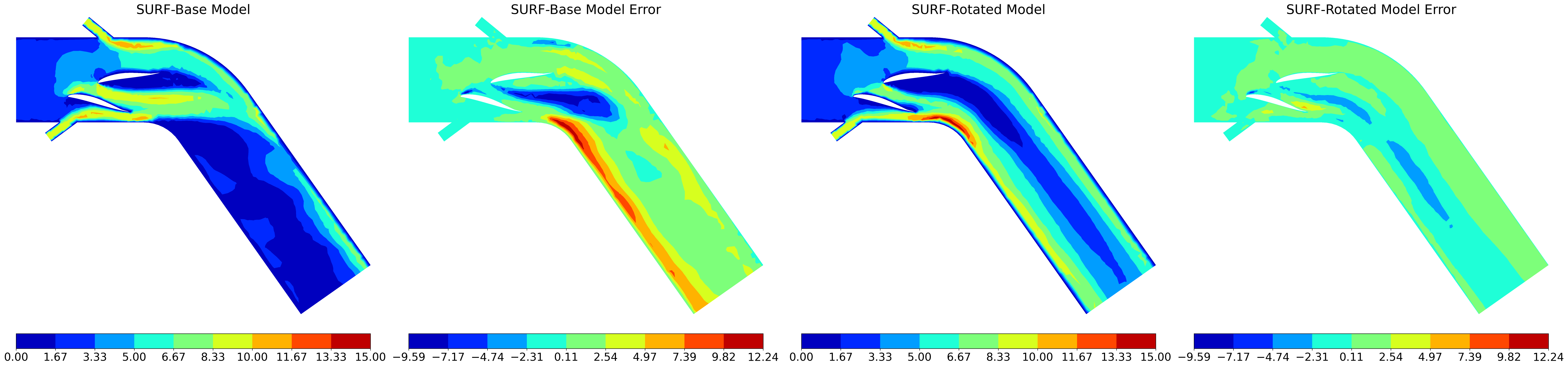}
	\caption{When training a model on the augmented dataset SURF-Rotated (right) instead of the SURF-Base dataset (left), the model exhibits better generalization to new environments and can even predict the flow between objects such as wing profiles more precisely, even though it has never encountered these during training. Values are displayed contoured to make visual comparisons easier.}
	\label{fig:turned}
\end{figure}

We reaffirm the conclusions drawn from the prior SURF generalization scores: EAGLE excels in generalization across all datasets, except Topology. Notably and maybe surprisingly, training on the SURF-Rotated dataset substantially improves generalization to the SURF-Full dataset for both MGN and EAGLE. Observing dynamic fluid boundary conditions modestly enhances generalization across datasets. Yet the most significant increase comes from the SURF-Rotated training. Recall that this dataset is a rotated version of the base dataset, and while training solely on rotated versions of the same dataset does not augment dataset size, introduce new topologies or dynamics, it remarkably amplifies generalization across all datasets. Illustrated in Figure \ref{fig:turned}, the model trained on rotated data excels in predicting fluid flow in adapted curved tunnels and amidst more challenging wing profiles. In contrast to a cylinder, a wing profile can redirect the flow.
Additionally, no flow separation occurs for low attack angles of an airfoil. With an increasing angle of attack, a flow separation may occur. Correctly predicting the flow separation or a stall is important for calculating the lift of an aircraft because a flow separation leads to a sudden loss of lift.

\section{Limitations}
The presented benchmark only considers incompressible flows, i.e., the fluid density is assumed to be constant over time and domain. This assumption is reasonable for flows where the velocity is less than a third of the speed of sound of the fluid. For dry air at a temperature of 20°C, the speed of sound is ~340 m/s. Hence, air flows with a speed of less than 100m/s can be considered incompressible. Further, the flow in the dataset is always turbulent; there is no laminar flow in the dataset because if laminar flows were included, the velocity range would become very large (>factor 1000). For these velocity ranges, the training is expected to be challenging. Additionally, the flow characteristics are different, further increasing the difficulty for a neural network to learn both fluid regimes.  

Another limitation presents itself in the fact that the datasets cannot capture all configurations that could show up in practice. While the presented metrics are well-suited to judge generalization capabilities to different domains, they can only serve as an approximation of how well these models generalize to real-world applications with variations that are not included in our testing. That being said, we expect that our metrics give a good and valuable indication of the generalization power of the tested model.
\paragraph{Future Work}
The proposed benchmark serves as a starting point for future research to test machine learning methods predicting fluid flows for generalization and overall performance. A possible future direction for the practical applications of learned neural networks on fluid flow predictions could be the application of these architectures to three-dimensional geometries. The biggest hurdle to overcome is the high memory footprint, especially during training. For example, the MGN method used approximately 24 GB of GPU memory during training. By simply extruding the geometry from the SURF-Full dataset by 100mm, the number of nodes in the mesh increases by a factor of $10$, therefore requiring considerably more memory. 
Thus, we face a new challenge for neural fluid simulators: Developing scalable architectures that can be applied to large (or high-resolution) and complex geometries.
Another interesting area for future investigations lies in exploring a broader spectrum of architectural choices and potential factors that can impact generalization. This encompasses the characteristics of the training, architecture choices, or even continual learning. SURF stands as a versatile platform well-suited for delving into this research.

\section{Conclusion}
Fluid simulation is a critical building block in developing and optimizing cutting-edge products, such as turbomachinery, cars, and airplanes. Using machine-learning models to simulate fluid dynamics can lead to faster processing times or more precise results. Given the nature of where such simulators are deployed, it is infeasible to train on all possible scenarios. Therefore, the ability of a learned simulator to grasp the underlying physics and generalize is an essential aspect of judging a given simulator's quality and applicability to real-world scenarios. We introduce SURF, a dedicated benchmark consisting of seven datasets to assess a mesh-based model's performance and generalization capabilities. We present a set of new metrics, which can be used to test and quantify generalization to different domain topologies, dynamic boundary conditions, finer mesh resolution, or a wider range of simulation parameter values.

We demonstrate the applicability of SURF on two state-of-the-art models and perform a comparative analysis. Both models achieve similar performance regarding the overall simulation accuracy. The main objective of SURF is to assess the generalization abilities. Here, the generalization scores show that while MGN generalizes better to unseen topologies, EAGLE can adapt better to the other aspects and achieves a lower unified $\text{SURF}_{\text{GS}}$ score. Interestingly, training on a rotated domain has a large positive impact on both models, an observation that can be used in future work to improve neural fluid simulators. 

Our benchmark focuses on how well learned mesh-based models adapt to new fluid dynamic domains, a crucial property for general-purpose fluid simulators. To achieve this, we introduce seven new large-scale datasets. Alongside these datasets, we introduce a new set of metrics: the SURF generalization scores. These metrics offer a simple way to measure different aspects of generalization and allow for easy comparison between different approaches. Our testing clearly showcases the effectiveness of SURF in evaluating generalization ability. As we move forward, we anticipate that SURF will contribute to a better understanding of generalization and a simpler and unified evaluation within this field.

\vfill

\bibliographystyle{unsrtnat}
\bibliography{reference}
\newpage
\appendix

\section{Extended Related Work} \label{sec:ExtendedRelatedWork}

The particle-based and grid-based methods for fluid simulation are described in the following.

\paragraph{Particle-Based Methods}
The particle-based approach to fluid simulations draws inspiration from smoothed particle hydrodynamics (SPH) as outlined in \cite{SPH}. SPH represents a mesh-free Lagrangian technique in which particle interactions serve as approximations of fluid dynamics. This approach proves advantageous in modeling situations like free surface flows (such as ocean waves or partially filled pipes) and intricate fluid-solid interactions (like water splashes resulting from a car moving through a water puddle). Notably, these scenarios involve fluid domains where considerable changes occur, even affecting domain boundaries throughout the simulation. This inherent adaptability simplifies the simulation process, eliminating the need for mesh adaptation due to the mesh-free nature of the method.
An example of the particle-based approach for fluid simulations with machine learning is ``Graph Network-based Simulators'' (GNS)~\cite{sanchezgonzalez2020learning}. The Graph Network framework~\cite{GraphNetworks} is applied, where particles are represented as nodes, and the interactions among the particles are modeled with edges. The dynamic behavior is replicated by learned message-passing on graphs. This approach was adapted by~\citet{LI2022201}, who represent the advection (viscosity and body forces), pressure, and collisions between particles through three separate graphs. 

\paragraph{Grid-Based Methods}
Grid-based fluid simulations are relatively fast and mainly used for real-time and visualization applications~\cite{Grid1, Grid2}. Leveraging the grid structure facilitates optimal GPU utilization. Nonetheless, a notable drawback is the consistent resolution across the domain, which may give rise to issues concerning precision and the portrayal of small entities within the flow. 
Due to the native GPU hardware support, many works on using machine learning methods to predict physics are based on convolutional architectures on regular grids~\cite{CNN1,CNN3, CNN4, CNN2}.

\section{Physical Background}
This section gives a brief background of the physics phenomena governing fluid dynamics, mesh generation, and how the quality and accuracy of a prediction are quantified. 
\paragraph{Fluid Simulation}
The physical behavior of fluids is described by the Navier-Stokes equations, a set of partial differential equations. They specifically express the conservation of mass and the momentum balance. In this work, we restrict ourselves to incompressible 2D flows with Newtonian fluids. 
The velocity and pressure field completely defines the flow of an incompressible fluid at a certain point in time. Two quantities are preserved in the fluid flow: mass and momentum. The equation for the conservation of mass for an incompressible, 2D flow is given by \cite{ansysHelpFluent}:
\begin{equation*}
    \frac{\partial u}{\partial x} + \frac{\partial v}{\partial y} = 0
\end{equation*}
Where $u$ and $v$ denote the velocity in the x-direction and y-direction, respectively. The momentum balance equations are:
\begin{align*}
    \rho \frac{\partial u}{\partial t} + \rho u \frac{\partial u}{\partial x} + \rho v \frac{\partial u}{\partial y} &= -\frac{\partial p}{\partial x} + \frac{\partial}{\partial x} \left[2\mu\frac{\partial u}{\partial x}\right] + \frac{\partial}{\partial y} \left[\mu\frac{\partial u}{\partial y} + \frac{\partial v}{\partial x}\right] \\
    \rho \frac{\partial v}{\partial t} + \rho u \frac{\partial v}{\partial x} + \rho v \frac{\partial v}{\partial y} &= -\frac{\partial p}{\partial y} + \frac{\partial}{\partial y} \left[2\mu\frac{\partial v}{\partial y}\right] + \frac{\partial}{\partial x} \left[\mu\frac{\partial u}{\partial y} + \frac{\partial v}{\partial x}\right] 
\end{align*}
$p$ denotes the pressure, whereas the fluid density and the kinematic viscosity are denoted by $\rho$ and $\mu$.
For the calculation of the temperature, the heat transfer has to be calculated. This leads to the energy conservation equation:
\begin{equation*}
    \rho c_p \left[\frac{\partial T}{\partial t} + u\frac{\partial T}{\partial x} + v\frac{\partial T}{\partial y} \right] = k\left[\frac{\partial^2 T}{\partial x^2} + \frac{\partial^2 T}{\partial y^2} \right]
\end{equation*}
$k$ and $c_p$ denote the thermal conductivity and heat capacity, respectively. If the fluid velocities $u$ and $v$ are set to zero, one gets the equation for the heat transfer in solids. Hence, the terms weighted with the fluid velocities describe the convective heat transport.

\paragraph{Reynolds Number} \label{Section:ReynoldsNumber}
Note that two incompressible flows with the same Reynolds number behave similarly and the velocity and pressure fields only differ by a constant factor. As a consequence, by scaling the solution, the simulation of a single fluid (e.g., water) can be translated to another medium (e.g., air).
The Reynolds number is an important dimensionless quantity used to characterize fluid flows. The Reynolds number is the ratio between inertial and viscous forces \cite{ReynoldsNumber}:
\begin{equation*}
    \text{Re} = \frac{\rho u L}{\mu}
\end{equation*}
$L$ is a characteristic linear dimension. For example, for a flow around a cylinder, the characteristic length is equal to the diameter of the cylinder, or for a flow in a pipe, it is equal to the diameter of the pipe. 

The Reynolds number is often used to predict the transition from laminar to turbulent flow. For a pipe flow with $\text{Re}<2000$, the flow is classified as laminar, for $\text{Re}>4000$, the flow is classified as turbulent \cite{ReynoldsPipe}. In between, the flow is classified as transitional. In laminar flows, the fluid particles follow smooth paths in layers. Particles close to a solid surface move in lines parallel to that surface. There are no swirls in the flow. In contrast, unsteady vortices across large spatial scales interact in turbulent flows.

\section{Dataset Generation}

\subsection{Parameters}
\begin{figure}
    \centering
    \includegraphics[width=1.0\columnwidth]{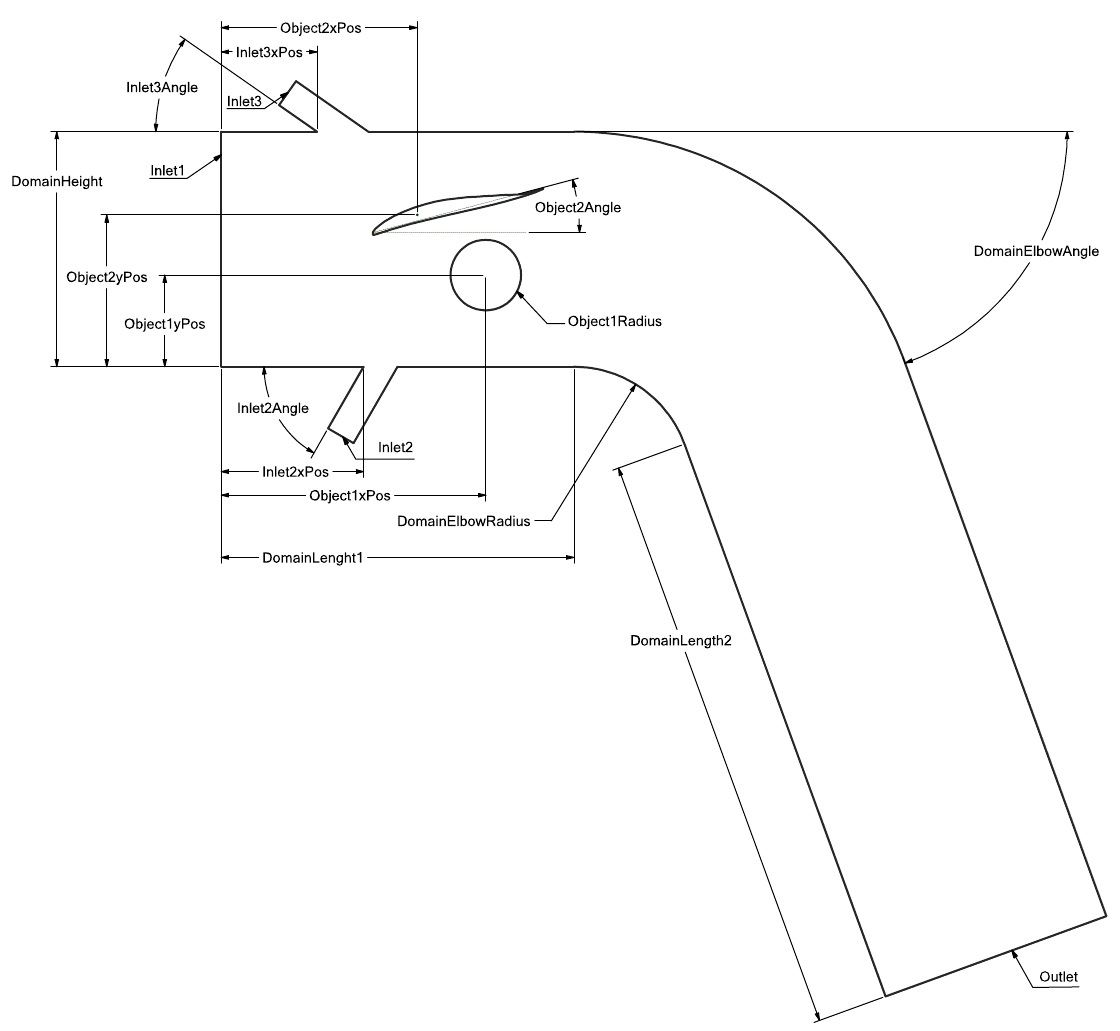}
    \caption{Sketch of how the individual geometry parameters influence the creation of the SURF dataset. Each dataset defines its own range for its generation parameter. For example, The SURF-Base dataset only allows for a cylinder while SURF-FULL can have up to two airfoils.}
    \label{fig:domain_parameters}
\end{figure}

The 2D geometry, in which the fluid flows, is built from the parameters shown in Figure~\ref{fig:domain_parameters}. Small gaps can lead to numerical problems during the solution (meshing and stability of the solver). To prevent too small gaps, the y-position of each object is calculated using a $ObjectyFactor$ ranging from 0 to 1. 
\begin{equation}
\begin{aligned}
    ObjectyPos = & minDistance + objectHeight + \\ 
    & ObjectyFactor \cdot (DomainHeight - 2 \cdot objectHeight \\
    & - 2 \cdot minDistance)
\end{aligned}
\end{equation}
Where $minDistance$ is set to 30mm and $objectHeight$ denotes the vertical size of the corresponding object.
The transient velocities of Inlet2 and Inlet3 are defined according to:
\begin{equation}
    Inletv(t) = InletvMean + InletvAmplitude \cdot \sin(2 \pi \cdot InletvFrequency \cdot t)
\end{equation}	
The outlet boundary condition does not allow an influx. Therefore, if necessary, the amplitude of Inlet2 and Inlet3 is reduced to always ensure a net outflow through the outlet. 

\begin{figure}
    \centering
    \includegraphics[width=0.6\columnwidth]{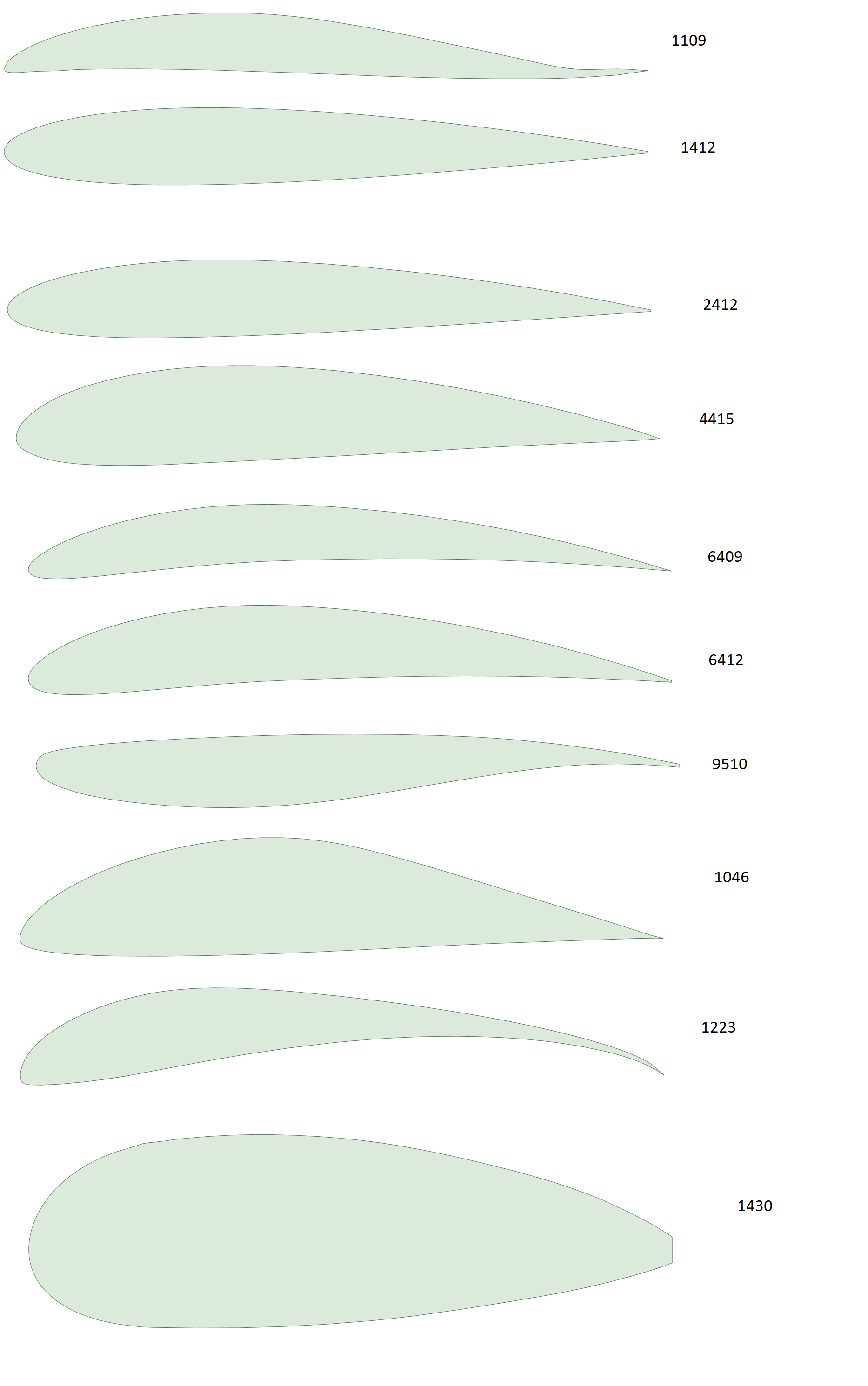}
    \caption{Different airfoil shapes and their corresponding number.}
    \label{fig:airfoils}
\end{figure}

The data sets and the used range for each parameter are given in Table~\ref{table:data_sets_geometry} and~\ref{table:data_sets_BC}. Given the range of the cylinder radius and inlet velocity, the SURF-Base and SURF-Base data set cover a Reynolds Number range between 1000 and 126000.
\begin{table}
    \centering
    \resizebox{\columnwidth}{!}{
    \begin{tabular}{ccccccc}
        \toprule
        \textbf{Parameter} & \textbf{Base} & \textbf{Rotated} & \textbf{Range} & \textbf{Topo} & \textbf{Dynamic} & \textbf{Full} \\ \midrule
        DomainLength & 1600 & 1600 & 1600 & 1600 & 1600 & 1600 \\ \midrule
        DomainHeight & 400 & 400 & 400 & 400 & 400 & 400 \\ \midrule
        DomainElbowAngle & 0 & 0 & 0 & [0, 90] & 0 & [0, 90] \\ \midrule
        DomainElbowRadius & 0 & 0 & 0 & 200 & 0 & 200 \\ \midrule
        DomainOrientation & 0 & [0, 360] & 0 & 0 & 0 & [0, 360] \\ \midrule
        Inlet2xPos & [150, 450] & [150, 450] & [150, 450] & [150, 450] & [150, 450] & [150, 450] \\ \midrule
        Inlet2Angle & [20, 45] & [20, 45] & [20, 45] & [20, 90] & [20, 45] & [20, 90] \\ \midrule
        Inlet3xPos & [150, 450] & [150, 450] & [150, 450] & [150, 450] & [150, 450] & [150, 450] \\ \midrule
        Inlet3Angle & [20, 45] & [20, 45] & [20, 45] & [20, 45] & [20, 45] & [20, 45] \\ \midrule
        Object1Type & Cylinder & Cylinder & Cylinder & Cylinder \& 5 Airfoils & Cylinder & Cylinder \& 10 Airfoils \\ \midrule
        Object1xPos & [150, 450] & [150, 450] & [150, 450] & [150, 450] & [150, 450] & [150, 450] \\ \midrule
        Object1yFactor & [0, 1] & [0, 1] & [0, 1] & [0, 1] & [0, 1] & [0, 1] \\ \midrule
        Object1Angle & n/a & n/a & n/a & [-15, 15] & n/a & [-30, 30] \\ \midrule
        Object1Radius & [45, 75] & [45, 75] & [30, 90] & [45, 75] & [45, 75] & [30, 90] \\ \midrule
        Object2Type & n/a & n/a & n/a & Cylinder \& 5 Airfoils & n/a & Cylinder \& 10 Airfoils \\ \midrule
        Object2xPos & n/a & n/a & n/a & [150, 450] & n/a & [150, 450] \\ \midrule
        Object2yFactor & n/a & n/a & n/a & [0, 1] & n/a & [0, 1] \\ \midrule
        Object2Angle & n/a & n/a & n/a & [-15, 15] & n/a & [-30, 30] \\ \midrule
        Object2Radius & n/a & n/a & n/a & [45, 75] & n/a & [45, 75] \\ \bottomrule
    \end{tabular}
    }
\caption{The geometry parameter ranges for the generation of the different SURF datasets. }
\label{table:data_sets_geometry} 
\end{table}

\begin{table}[!ht]
    \centering
    \resizebox{\columnwidth}{!}{
    \begin{tabular}{ccccccc}
        \toprule
        \textbf{Parameter} & \textbf{Base} & \textbf{Rotated} & \textbf{Range} & \textbf{Topo} & \textbf{Dynamic}  & \textbf{Full} \\ \midrule
        Inlet1v & [1, 10] & [1, 10] & [0.5, 20] & [1, 10] & [1, 10] & [0.5, 20] \\ \midrule
        Inlet2vMean & [1, 10] & [1, 10] & [1, 10] & [1, 10] & [1, 10] & [1, 10] \\ \midrule
        Inlet2vAmplitude & 0 & 0 & 0 & 0 & [0, 10] & [0, 10] \\ \midrule
        Inlet2vFrequency & 0 & 0 & 0 & 0 & [1, 5] & [1, 5] \\ \midrule
        Inlet2T & [290, 310] & [290, 310] & [290, 310] & [290, 310] & [290, 310] & [290, 310] \\ \midrule
        Inlet3vMean & [1, 10] & [1, 10] & [1, 10] & [1, 10] & [1, 10] & [1, 10] \\ \midrule
        Inlet3vAmplitude & 0 & 0 & 0 & 0 & [0, 10] & [0, 10] \\ \midrule
        Inlet3vFrequency & 0 & 0 & 0 & 0 & [1, 5] & [1, 5] \\ \midrule
        Inlet3T & [290, 310] & [290, 310] & [290, 310] & [290, 310] & [290, 310] & [290, 310] \\ \midrule
        Object1T & [450, 800] & [450, 800] & [375, 1300] & [450, 800] & [450, 800] & [375, 1300] \\ \midrule
        Object2T & n/a & n/a & n/a & [450, 800] & n/a & [375, 1300] \\ \midrule
        Thermal Conductivity & [0.0258, 0603] & [0.0258, 0603] & [0.013, 1.2] & [0.0258, 0603] & [0.0258, 0603] & [0.013, 1.2] \\ \midrule
        Heat Capacity & [1.02, 3223] & [1.02, 3223] & [0.5, 6446] & [1.02, 3223] & [1.02, 3223] & [0.5, 6446] \\ \bottomrule
    \end{tabular}
    }
\caption{The geometry parameter ranges for the generation of the different SURF datasets.}
\label{table:data_sets_BC} 
\end{table}

\subsection{Process for Generation of Data Sets}
The process of generating a data set is outlined in Figure~\ref{fig:mesh}. The steps required and the software used in the process are:
\begin{itemize}
    \item \textbf{Parameter Ranges:} Define the range of each parameter (see Table \ref{table:data_sets_BC} and \ref{table:data_sets_geometry}).
    \item \textbf{Latin Hypercube Sampling:} According to the defined range, Latin hypercube sampling \cite{LatinHypercube} is used to generate the samples for the data set. Each sample corresponds to a simulation with a specific geometry and boundary conditions defined by the parameters. These samples are internally referred to as design points (DP). For a sample size of $n$ the Latin hypercube algorithm divides each parameter range into $n$ equally long sub-ranges and ensures that there is only one sample in each sub-range. For two parameters, this is similar to having $n$ rooks on a $nxn$ chessboard without threatening each other. \newline
    \item \textbf{Create Geometry \& Mesh:} The geometry is built running in Ansys SpaceClaim 2023 R1 \cite{ansysSpaceClaim2023R1}. The creation of the geometry and the mesh for each design point is done in Ansys Workbench 2023 R1 \cite{ansysWorkbench2023R1}, and Ansys Meshing 2023 R1 \cite{ansysMeshing2023R1}. Two different meshes are created with Ansys Meshing\cite{ansysMeshing2023R1}. The fine mesh consists mostly of quadrilaterals and a small amount of triangles. Quadrilaterals are preferred because they are numerically more efficient (better solution quality with the same cost compared to triangles). At the wall boundaries, an inflation layer with eight elements is defined for a better gradient resolution. This mesh is used for the ground truth calculation. The coarse mesh, consisting of triangles, is used for training and predictions (cf. Figure~\ref{fig:mesh}). A coarse mesh must be used due to the GPU memory footprint of the training and to ensure that the training can still be run on an Nvidia RTX 3090 GPU (24 GB). \newline
    Software: Ansys SpaceClaim 2023 R1, Ansys Workbench 2023 R1, Ansys Meshing 2023 R1
    \item \textbf{Prepare for execution:} The fluid simulation setup is defined in a journal file for Ansys Fluent (FluentSimulation.jou). In this file, the simulation settings, boundary conditions, and fluid parameters are defined. We then run the simulation on a compute cluster.
    \item \textbf{Run:} Each simulation is run as a separate job, requesting two cores and 2 GB of memory. 
    A transient incompressible 2D simulation is run with a fixed time step of 0.01s for 300 time steps. The calculated results on the fine mesh are then mapped to the coarse mesh. For the downsampling linear triangular interpolation is used. Due to numerical inaccuracies, some nodes lie outside of the domain of the mesh. For these nodes, the nearest neighbor interpolation is used. The final data structure for the training data set was taken from the Eagle paper. For each DP there is a folder with the two files sim.npz and triangles.npy. sim.npz contains the node coordinates, velocities, pressure, and temperature for each calculated time step. Triangles.npy contains the edges connecting the nodes. \newline
    Software: Python 3.8.5 and Ansys Fluent 2021 R1
\end{itemize}

\subsection{Dataset Statistics}
 Before downsampling, each data set has a size of about 2 TB. The size of the downsampled data sets used for the training is between 37 and 53 GB. The calculation of one simulation takes between 7 - 30min to finish. Each dataset is divided into a training, validation, and test set, by randomly shuffling the simulations and selecting 80\% of the simulations for the training set and 10\% for the validation and training set respectively. Table~\ref{table:FullStatistics} shows statistics for the main parameters of the generated dataset. Key characterics of the datasets are given in Table~\ref{table:data_sets_overview}.

 \begin{table}
    \centering
    \resizebox{\columnwidth}{!}{
    \begin{tabular}{cccccccc}
    \toprule
         & \textbf{Base} & \textbf{Rotated} & \textbf{Range} & \textbf{Topo} & \textbf{Dynamic}  & \textbf{Full} & \textbf{Full Finer} \\ \midrule
        Number of Datapoints & 1200 & 1200 & 1200 & 1315 & 1200 & 1241 & 1241\\ \midrule
        Time Steps & 300 & 300 & 300 & 300 & 300 & 300 & 300\\ \midrule
        End Time [s] & 3 & 3 & 3 & 3 & 3 & 3 & 3\\ \midrule
        Average Number of Nodes & 1287 & 1288 & 1293 & 1751 & 1287 & 1787 & 5433\\ \midrule
        Dataset Size [GB] & 10.9 & 10.8 & 10.9 & 16.0 & 11.3 & 16.7 & 68.3\\ \midrule
    \end{tabular}
    }
\caption{Overview of the different SURF datasets.}
\label{table:data_sets_overview} 
\end{table}

 \begin{table}
    \caption{Statistics for the state variables of the SURF-Full data set.}
    \centering
    \begin{tabular}{ccc}
        \toprule
        ~ & \textbf{Mean} & \textbf{Standard Deviation} \\ \midrule
        Velocity & -1 & 9 \\ \midrule
        Pressure & 59 & 137 \\ \midrule
        Temperature & 340 & 129 \\ \bottomrule
    \end{tabular}
    \label{table:FullStatistics}
\end{table}

\section{Baselines} \label{chapter:Architecture}
Two Graph Neural Network models are trained on the presented data sets. Because the training data also includes the temperature as a state value, the models are slightly adapted to predict the temperature field in the flow as well. 

\subsection{Training and Loss} \label{sec:app_training}
Both models are trained with seven-step supervision. The predictions are calculated as follows:
\begin{equation}
\begin{aligned}
    \textbf{q}_P^{t+1} &= \textbf{q}_{GT}^{t} + \textbf{p}^1 \\
    \textbf{q}_P^{t+2} &= \textbf{q}_P^{t+1} + \textbf{p}^2 \\
    &\vdots \\
    \textbf{q}_P^{t+7} &= \textbf{q}_P^{t+6} + \textbf{p}^6 \\
\end{aligned}
\end{equation}
The loss is calculated as:
\begin{equation}
    loss = \text{MSE} \left(target, output \right) \\
    \label{eq:EAGLELoss}
\end{equation}
with:
\begin{equation}
    target = \begin{bmatrix} \textbf{q}_{GT}^{t+1} - \textbf{q}_{GT}^t \\ \textbf{q}_{GT}^{t+2} - \textbf{q}_{P}^{t+1} \\ \vdots \\ \textbf{q}_{GT}^{t+7} - \textbf{q}_{P}^{t+6} \end{bmatrix} , output = \begin{bmatrix} \textbf{p}^1 \\ \vdots \\ \textbf{p}^7\end{bmatrix}
\end{equation}
$\textbf{q}^{t}$ denotes the state at time step $t$ the subscripts $GT$ and $P$ are used to specify whether the state values are from the ground truth or predicted values, respectively. Note that the $\text{target}$ values are always calculated with respect to the previously predicted state and not the state from the ground truth.

The authors of EAGLE show that both models are more robust if trained on more than one time step. Hence, for better comparability, we train both models with seven-step supervision. We use the implementation provided by EAGLE~\cite{EagleGit} for both models. In this implementation, the loss for the MGN model is calculated differently.
In contrast to the loss for the EAGLE model, the $target_{MGN}$ is calculated solely on the ground truth data and is independent of the predicted data. We have found that using the loss given in Eq.~\ref{eq:EAGLELoss} gives better results for the MGN model and therefore use the same loss for both models. This also lets us focus more on differences in the architectures.

\begin{figure}
    \centering
    \includegraphics[width=0.8\linewidth]{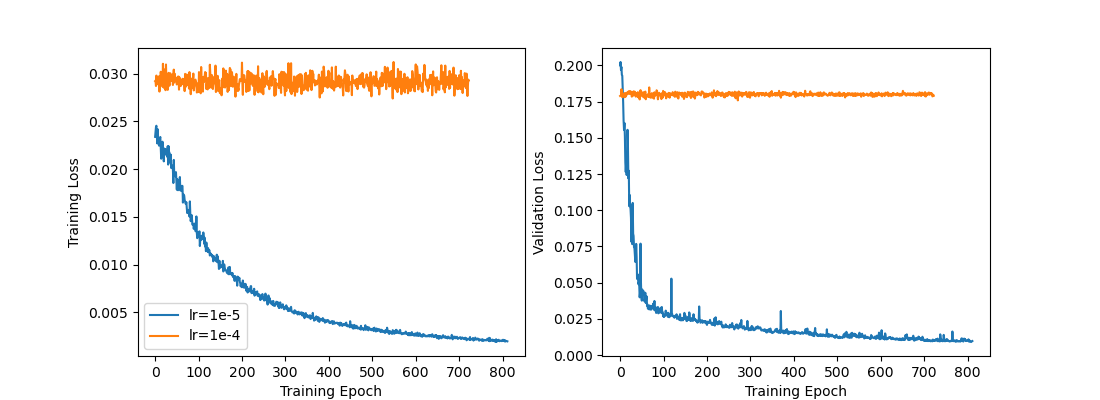}
    \caption{Training and validation loss for the EAGLE model with different training parameters for the SURF-Full data set.}
    \label{fig:GV_TrainingValidationLoss}
\end{figure}

The learning rate of $10^{-4}$ is used in the EAGLE paper~\cite{EaglePaper}. As shown in Fig.~\ref{fig:GV_TrainingValidationLoss} for the SURF-Full data set, the training works better with a learning rate of $10^{-5}$. We therefore use a learning rate of $10^{-5}$ in our experiments. 
For MGN we use a learning rate of $10^{-4}$ with a decay factor of $0.999$.

\subsection{Heat Transfer}
In contrast to the data set used in the MGN and EAGLE paper, the SURF datasets include heat transfer and therefore the temperature as a state variable. Hence, the node encoders and decoders are adapted to include the temperature in the state as input value for the node features and output value, respectively. Apart from the flow velocity, two fluid properties characterize heat transfer: heat capacity and thermal conductivity. In Table~\ref{table:data_sets_BC}, the range of the two properties used to calculate the data sets is reported. These two properties are fed to the Neural Networks as additional node features to each node. Hence, the node features consist of a concatenation of the following: state (velocity, pressure, temperature), one hot encoding of the node type, and the thermal fluid properties.

\subsection{Computational Resources}
Trainings were executed on Nvidia Geforce RTX 3090 GPU's with a memory of 24 GB. The MGN model required 24 GB of memory for the SURF-Full data set, where the number of nodes is the highest, while the EAGLE model used 12 GB GPU memory. The 1,000 epochs of the EAGLE model took around 32h. The training for the MGN models was aborted after 48h reaching 700 - 800 epochs. The training for SURF-Mesh required more GPU memory and was done on A6000s with 48 GB of GPU memory for the EAGLE baseline and on A100 80GB GPU memory for the MGN baseline respectively.

\section{Full Empirical Evaluation}
We report the full performance and generalization scores in Table~\ref{table:PS_complete} and \ref{table:GS_Combined_Complete}:
\begin{table}
    \caption{Performance scores for velocity, pressure, and temperature after 250 times steps of MGN and EAGLE trained and tested on all datasets.}
    \centering
    \begin{tabular}{cccccc}
    \toprule
        \textbf{Dataset} & \textbf{Model} & \textbf{PS$_\text{v}$ $\downarrow$} & \textbf{PS$_\text{p}$ $\downarrow$} & \textbf{PS$_\text{t}$ $\downarrow$} & \textbf{PS  $\downarrow$} \\
        \midrule
        Base & MGN & 0.073 $\pm$ 0.005 &  0.142 $\pm$ 0.011 &  0.068 $\pm$ 0.003 & 0.094 $\pm$ 0.006    \\ \cmidrule{2-6}
             & EAGLE & 0.074 $\pm$ 0.003 & 0.090 $\pm$ 0.004 &  0.097 $\pm$ 0.006 & 0.087 $\pm$ 0.004   \\ \cmidrule{1-6}
        Rotated & MGN & 0.041 $\pm$ 0.002 &  0.133 $\pm$ 0.003 & 0.071 $\pm$0.002 & 0.082 $\pm$ 0.001   \\ \cmidrule{2-6}
               & EAGLE & 0.061 $\pm$ 0.007 & 0.089 $\pm$ 0.005 &  0.107 $\pm$ 0.003 & 0.086 $\pm$ 0.004 \\ \cmidrule{1-6}
		Topology & MGN & 0.094 $\pm$ 0.019 & 0.155 $\pm$ 0.037 & 0.087 $\pm$ 0.008 & 0.112 $\pm$ 0.021  \\ \cmidrule{2-6}
		        & EAGLE & 0.087 $\pm$ 0.006 & 0.092 $\pm$ 0.003 & 0.110 $\pm$ 0.007 & 0.097 $\pm$ 0.002 \\ \cmidrule{1-6}
		Range & MGN & 0.088 $\pm$ 0.004 & 0.130 $\pm$ 0.005 & 0.070 $\pm$ 0.010 & 0.096 $\pm$ 0.005     \\ \cmidrule{2-6}
		     & EAGLE & 0.111 $\pm$ 0.012 & 0.121 $\pm$ 0.015 & 0.084 $\pm$ 0.004 & 0.106 $\pm$ 0.008    \\ \cmidrule{1-6}
		Dynamic & MGN & 0.045 $\pm$ 0.008 & 0.050 $\pm$ 0.004 & 0.035 $\pm$ 0.002 & 0.043 $\pm$ 0.004   \\ \cmidrule{2-6}
		     & EAGLE  & 0.131 $\pm$ 0.020 & 0.123 $\pm$ 0.008 & 0.109 $\pm$ 0.008 & 0.121 $\pm$ 0.011   \\ \cmidrule{1-6}
        Full & MGN & 0.165 $\pm$ 0.099  &  0.168 $\pm$ 0.088 &  0.111 $\pm$ 0.041 & 0.148 $\pm$ 0.076   \\ \cmidrule{2-6}
        & EAGLE & 0.163 $\pm$ 0.004&  0.166 $\pm$ 0.011 &  0.146 $\pm$ 0.015 & 0.158 $\pm$ 0.006        \\ \cmidrule{1-6}
		Mesh & MGN & 0.292 $\pm$ 0.013 & 0.294 $\pm$ 0.028 & 0.211 $\pm$ 0.012 & 0.266 $\pm$ 0.017 \\ \cmidrule{2-6}
		         & EAGLE & 0.237 $\pm$ 0.010 & 0.248 $\pm$ 0.017 & 0.237 $\pm$ 0.016 & 0.240 $\pm$ 0.007 \\
        \bottomrule
    \end{tabular}
    \label{table:PS_complete}
\end{table}

\begin{table}
\caption{Generalization scores for 250 time steps the MGN and EAGLE model trained on all training sets and then tested on the test set SURF-Base, SURF-Rotated, SURF-Topology, SURF-Range, SURF-Dynamic, SURF-Full, and SURF-Mesh. Note that this is different from the SURF generalization scores.}
    \centering
    \resizebox{\columnwidth}{!}{
    \begin{tabular}{ccccccccc}
        \toprule
        \textbf{Training Set} & \textbf{Model} & \textbf{Base $\downarrow$} & \textbf{Rotated $\downarrow$} & \textbf{Topology $\downarrow$} & \textbf{Range $\downarrow$} & \textbf{Dynamic $\downarrow$} & \textbf{Full $\downarrow$} & \textbf{Mesh $\downarrow$}\\ \midrule
        Base & MGN & 1.00 $\pm$ 0.00 & 9.16 $\pm$ 0.54 & 3.10 $\pm$ 0.72 & 1.30 $\pm$ 0.07 & 5.76 $\pm$ 0.64 & 3.32 $\pm$ 1.30 & 1.55 $\pm$ 0.22	\\ \cmidrule{2-9}
        & EAGLE & 1.00 $\pm$ 0.00 & 6.67 $\pm$ 1.44 & 3.68 $\pm$ 0.10 & 1.08 $\pm$ 0.09 & 2.17 $\pm$ 0.08 & 3.73 $\pm$ 0.52 & 2.33 $\pm$ 0.14      \\ \cmidrule{1-9}
        Rotated & MGN & 1.08 $\pm$ 0.07 & 1.00 $\pm$ 0.00 & 1.81 $\pm$ 0.38 & 1.54 $\pm$ 0.18 & 6.73 $\pm$ 0.60 & 1.95 $\pm$ 0.72 & 1.09 $\pm$ 0.11      \\ \cmidrule{2-9}
        & EAGLE & 1.63 $\pm$ 0.13 & 1.00 $\pm$ 0.00 & 1.96 $\pm$ 0.08 & 1.36 $\pm$ 0.22 & 2.41 $\pm$ 0.17 & 1.47 $\pm$ 0.06 & 1.04 $\pm$ 0.05    \\ \cmidrule{1-9}
	Topology & MGN & 1.52 $\pm$ 0.34 & 7.81 $\pm$ 0.54 & 1.00 $\pm$ 0.00 & 1.48 $\pm$ 0.26 & 7.86 $\pm$ 0.92 & 2.96 $\pm$ 1.24 & 1.50 $\pm$ 0.04    \\ \cmidrule{2-9}
	& EAGLE & 2.49 $\pm$ 0.11 & 4.81 $\pm$ 0.28 & 1.00 $\pm$ 0.00 & 1.89 $\pm$ 0.10 & 3.27 $\pm$ 0.26 & 3.07 $\pm$ 0.26 & 2.22 $\pm$ 0.30     \\ \cmidrule{1-9}
	Range & MGN & 1.94 $\pm$ 0.27 & 11.95 $\pm$ 1.63 & 3.79 $\pm$ 0.96 & 1.00 $\pm$ 0.00 & 7.63 $\pm$ 0.75 & 4.21 $\pm$ 1.84 & 2.36 $\pm$ 0.73 \\ \cmidrule{2-9}
	& EAGLE & 1.89 $\pm$ 0.06 & 8.24 $\pm$ 2.08 & 3.60 $\pm$ 0.03 & 1.00 $\pm$ 0.00 & 2.33 $\pm$ 0.20 & 4.81 $\pm$ 2.20 & 2.81 $\pm$ 1.11 \\ \cmidrule{1-9}
       Dynamic & MGN & 0.58 $\pm$ 0.03 & 9.09 $\pm$ 0.35 & 3.11 $\pm$ 0.56 & 1.52 $\pm$ 0.23 & 1.00 $\pm$ 0.00 & 3.93 $\pm$ 1.52 & 1.94 $\pm$ 0.16 \\ \cmidrule{2-9}
        & EAGLE & 2.14 $\pm$ 0.19 & 10.96 $\pm$ 2.49 & 4.47 $\pm$ 0.24 & 2.16 $\pm$ 0.33 & 1.00 $\pm$ 0.00 & 5.25 $\pm$ 1.76 & 3.99 $\pm$ 1.58 \\ \cmidrule{1-9}
	Full & MGN & 3.53 $\pm$ 2.10 & 3.51 $\pm$ 2.26 & 1.98 $\pm$ 0.76 & 2.41 $\pm$ 1.42 & 6.30 $\pm$ 3.08 & 1.00 $\pm$ 0.00 & 1.07 $\pm$ 0.26      \\ \cmidrule{2-9}
	& EAGLE Full & 5.22 $\pm$ 0.49 & 2.58 $\pm$ 0.13 & 3.50 $\pm$ 0.45 & 4.04 $\pm$ 0.94 & 3.96 $\pm$ 0.38 & 1.00 $\pm$ 0.00 & 1.01 $\pm$ 0.05 \\ \cmidrule{1-9}
	Mesh & MGN  & 4.63 $\pm$ 0.42 & 4.50 $\pm$ 0.71 & 3.07 $\pm$ 0.97 & 2.91 $\pm$ 0.33 & 9.03 $\pm$ 1.41 & 1.70 $\pm$ 0.71 & 1.00 $\pm$ 0.00 \\ \cmidrule{2-9}
	& EAGLE & 5.07 $\pm$ 0.38 & 3.63 $\pm$ 0.23 & 3.97 $\pm$ 0.16 & 4.04 $\pm$ 0.88 & 3.80 $\pm$ 0.37 & 1.71 $\pm$ 0.10 & 1.00 $\pm$ 0.00 \\
        \bottomrule
    \end{tabular}
    }
    
    \label{table:GS_Combined_Complete}
\end{table}

\section{Examples}

\begin{figure}
    \centering
    \includegraphics[width=0.8\linewidth]{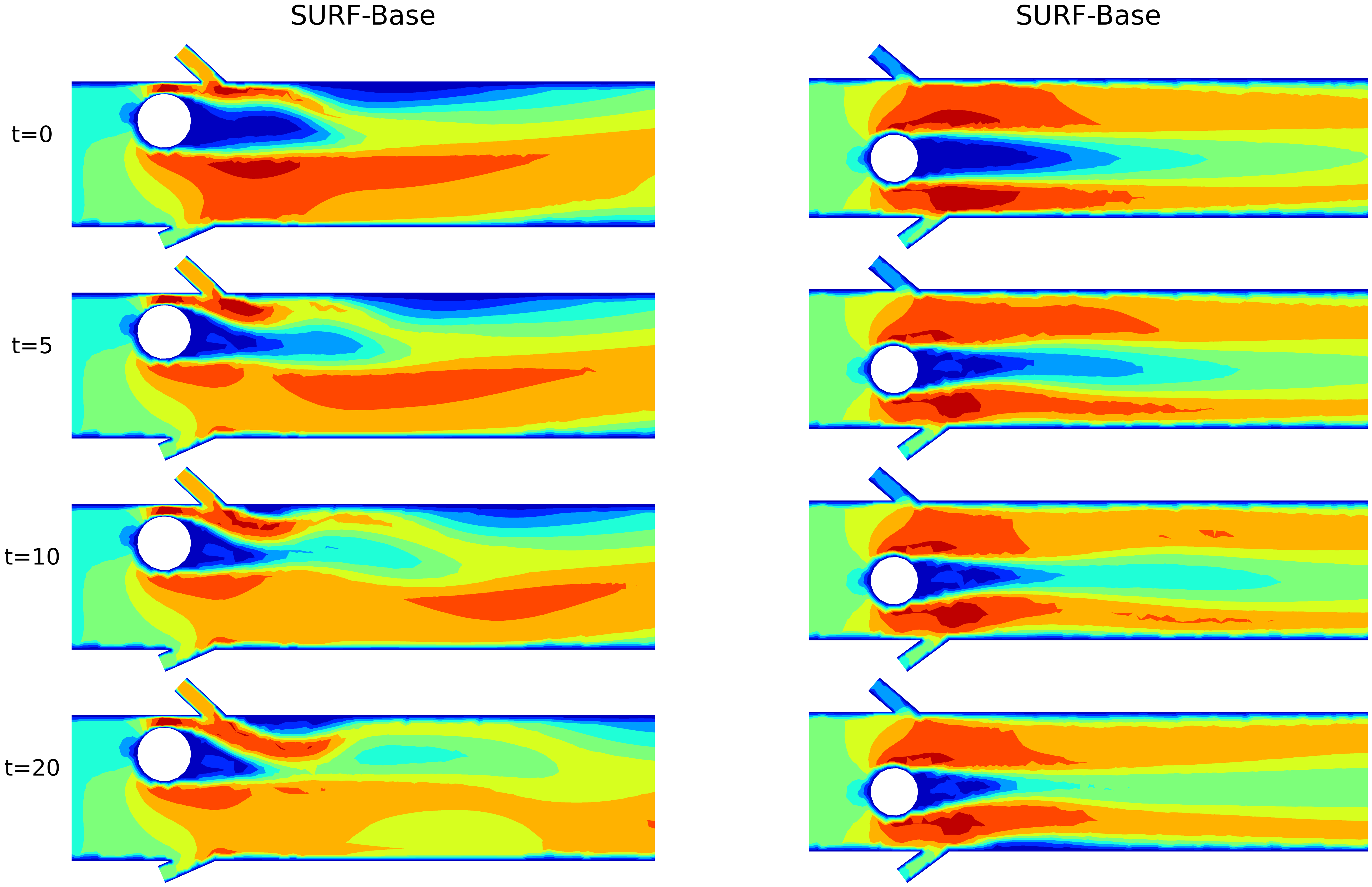}
    \caption{Velocity visualization of two ground-truth datapoints from SURF-Base, evaluated at timesteps 0, 5, 10, 20 (top to bottom).}
    \label{fig:BaseDatasetExamples}
\end{figure}

\begin{figure}
    \centering
    \includegraphics[width=0.8\linewidth]{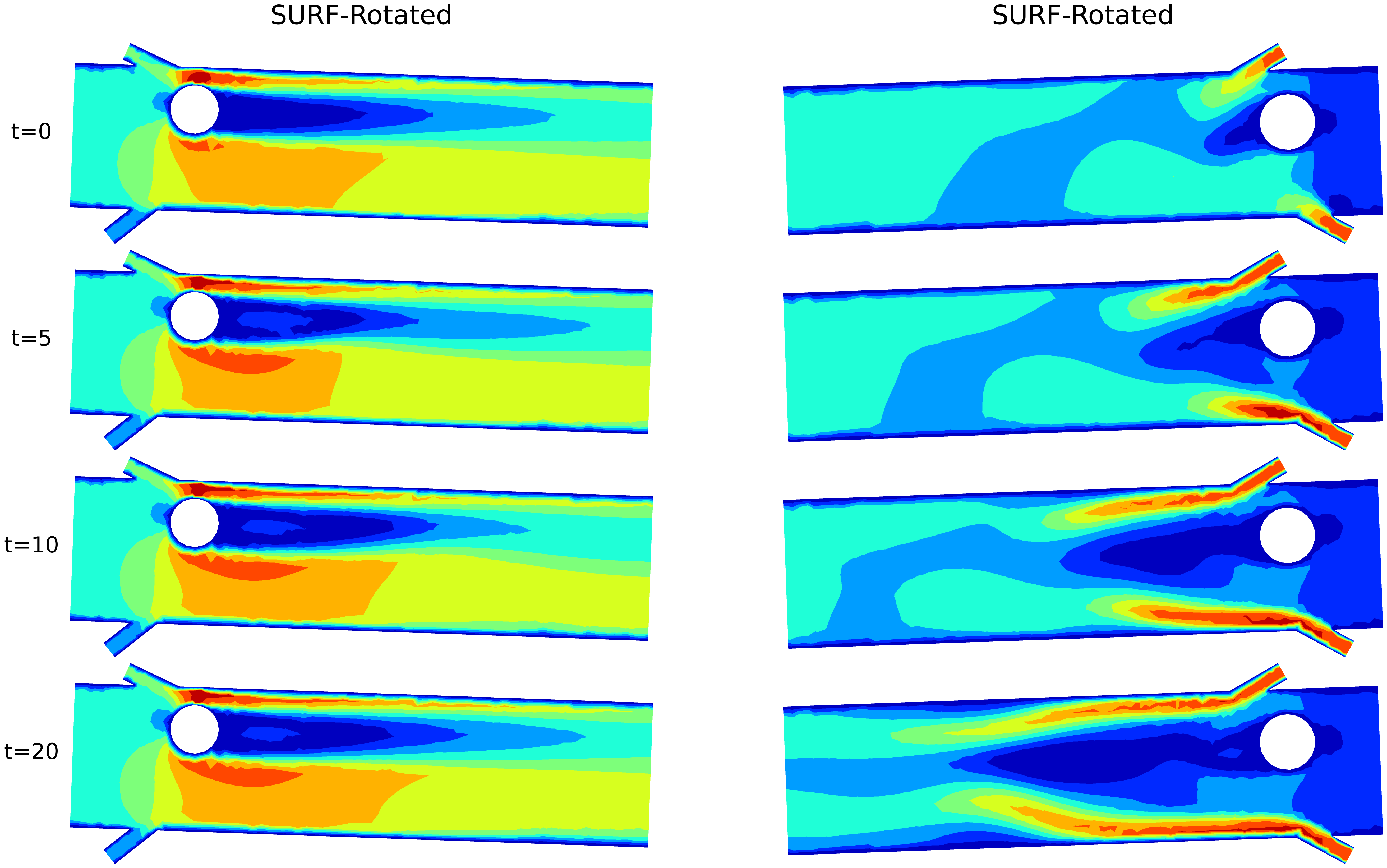}
    \caption{Velocity visualization of two ground-truth datapoints from SURF-Rotated, evaluated at timesteps 0, 5, 10, 20 (top to bottom).}
    \label{fig:RotatedDatasetExamples}
\end{figure}

\begin{figure}
    \centering
    \includegraphics[width=0.8\linewidth]{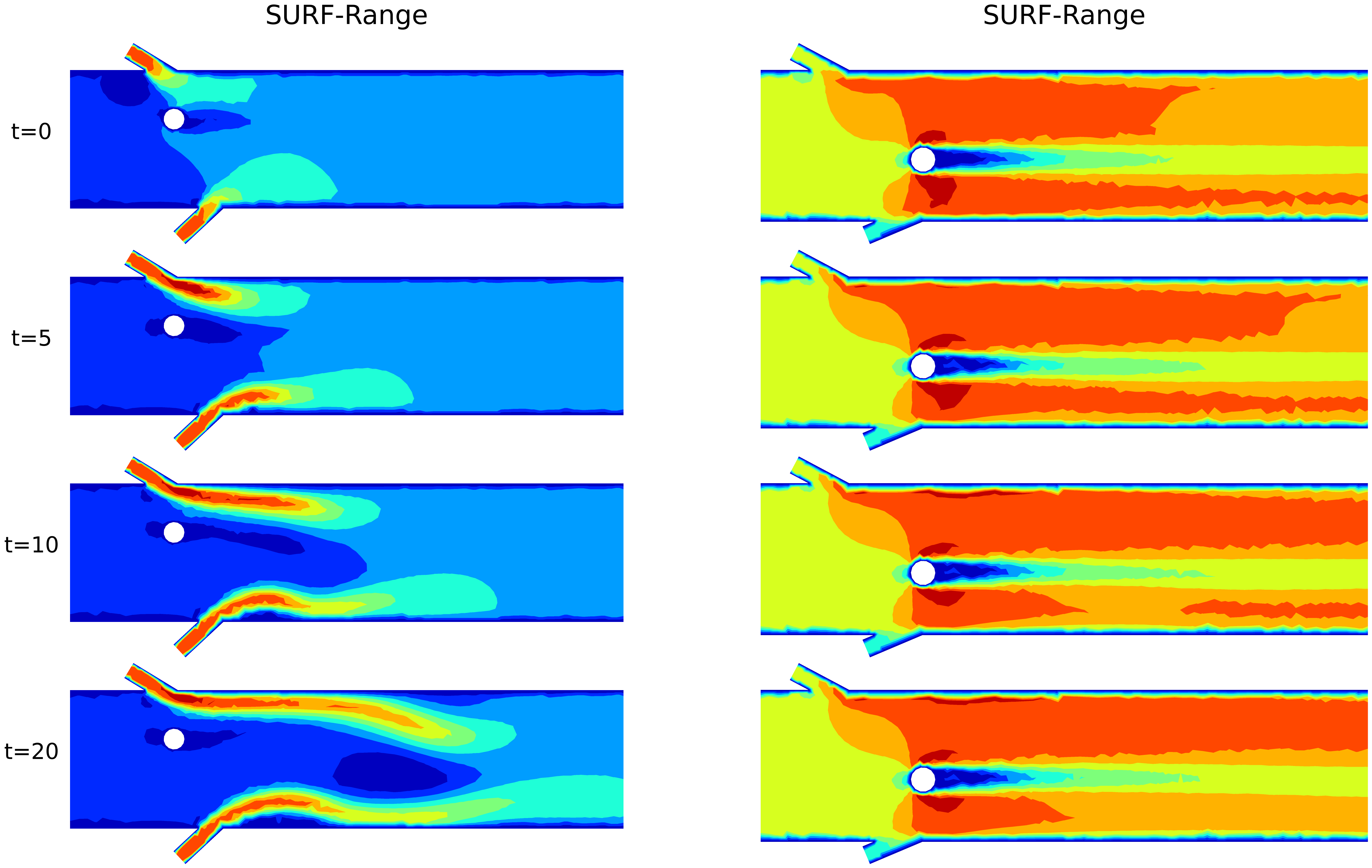}
    \caption{Velocity visualization of two ground-truth datapoints from SURF-Range, evaluated at timesteps 0, 5, 10, 20 (top to bottom).}
    \label{fig:RangeDatasetExamples}
\end{figure}

\begin{figure}
    \centering
    \includegraphics[width=0.8\linewidth]{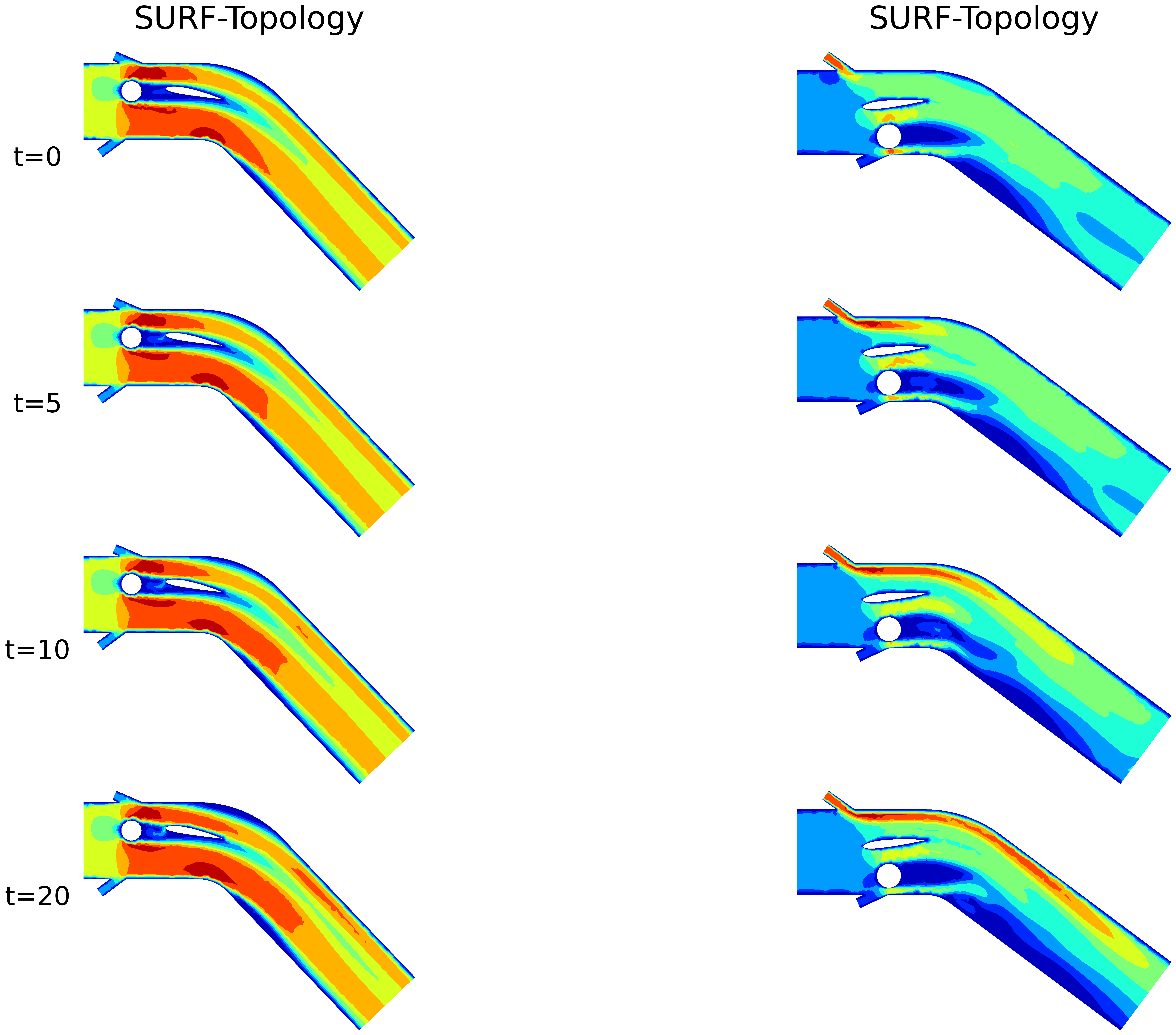}
    \caption{Velocity visualization of two ground-truth datapoints from SURF-Topology, evaluated at timesteps 0, 5, 10, 20 (top to bottom).}
    \label{fig:TopologyDatasetExamples}
\end{figure}

\begin{figure}
    \centering
    \includegraphics[width=0.8\linewidth]{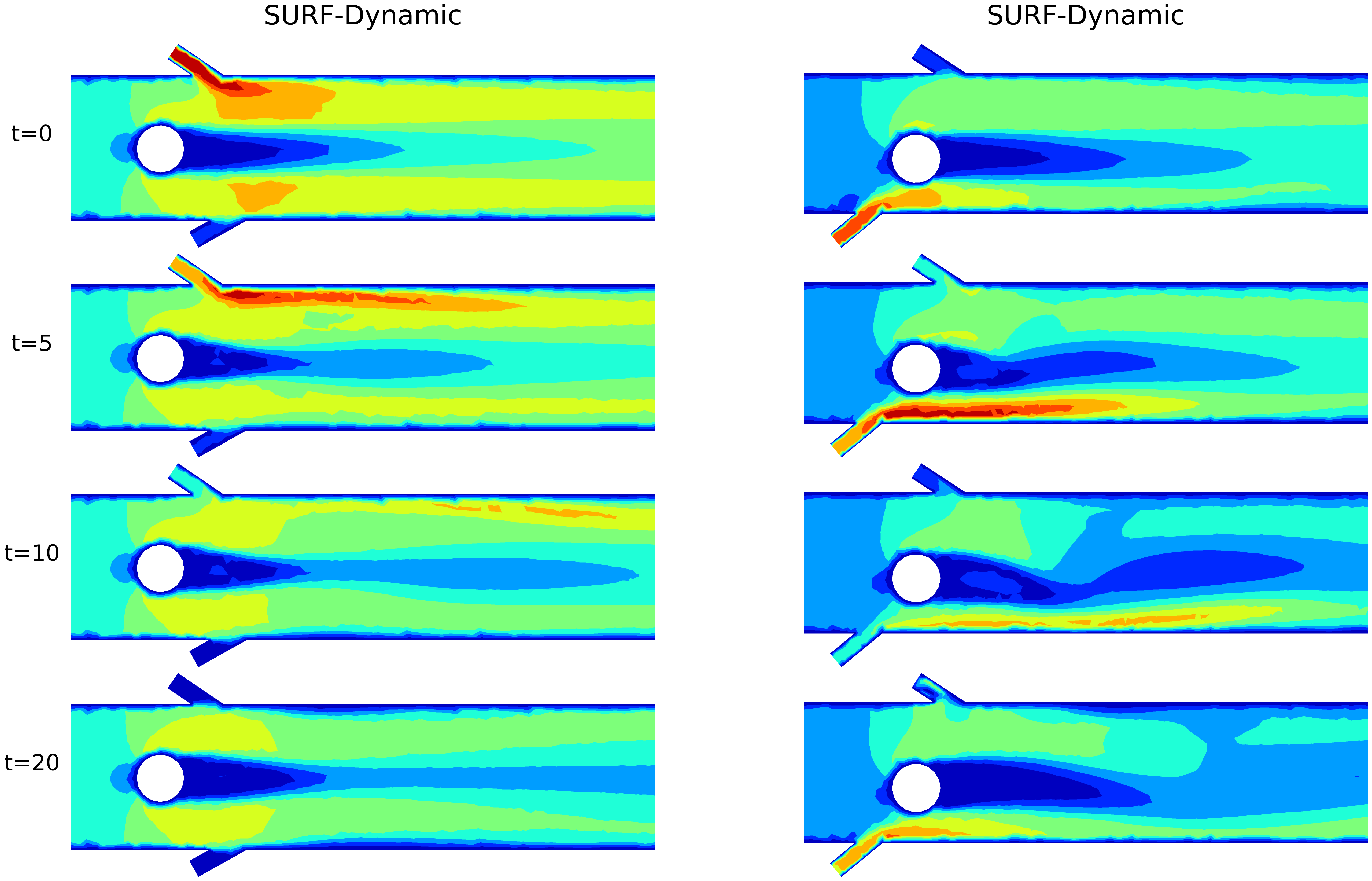}
    \caption{Velocity visualization of two ground-truth datapoints from SURF-Dynamic, evaluated at timesteps 0, 5, 10, and 20 (top to bottom).}
    \label{fig:DynamicDatasetExamples}
\end{figure}

\begin{figure}
    \centering
    \includegraphics[width=0.8\linewidth]{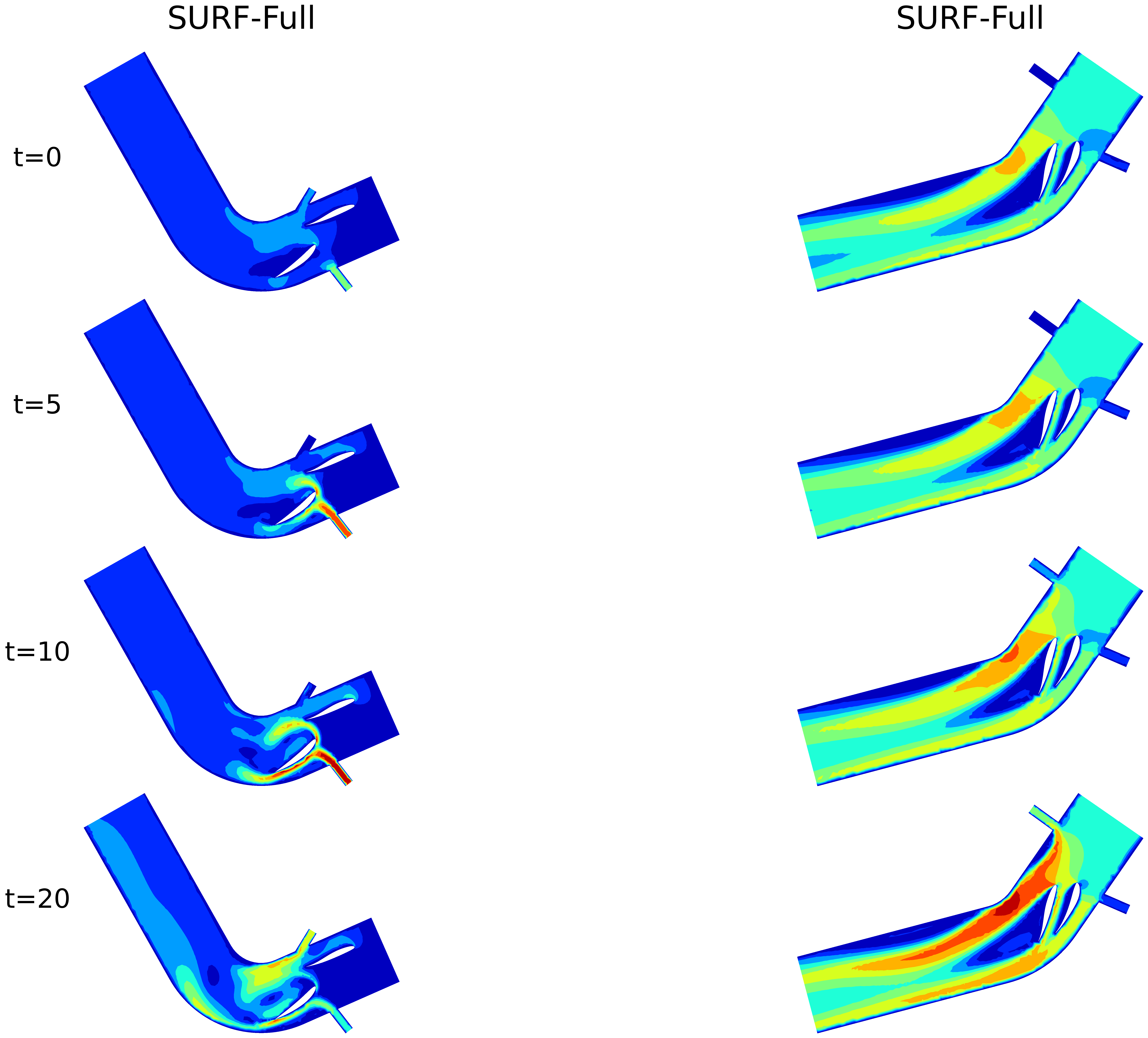}
    \caption{Velocity visualization of two ground-truth datapoints from SURF-Full, evaluated at timesteps 0, 5, 10, 20 (top to bottom).}
    \label{fig:FullDatasetExamples}
\end{figure}

\begin{figure}
    \centering
    \includegraphics[width=0.8\linewidth]{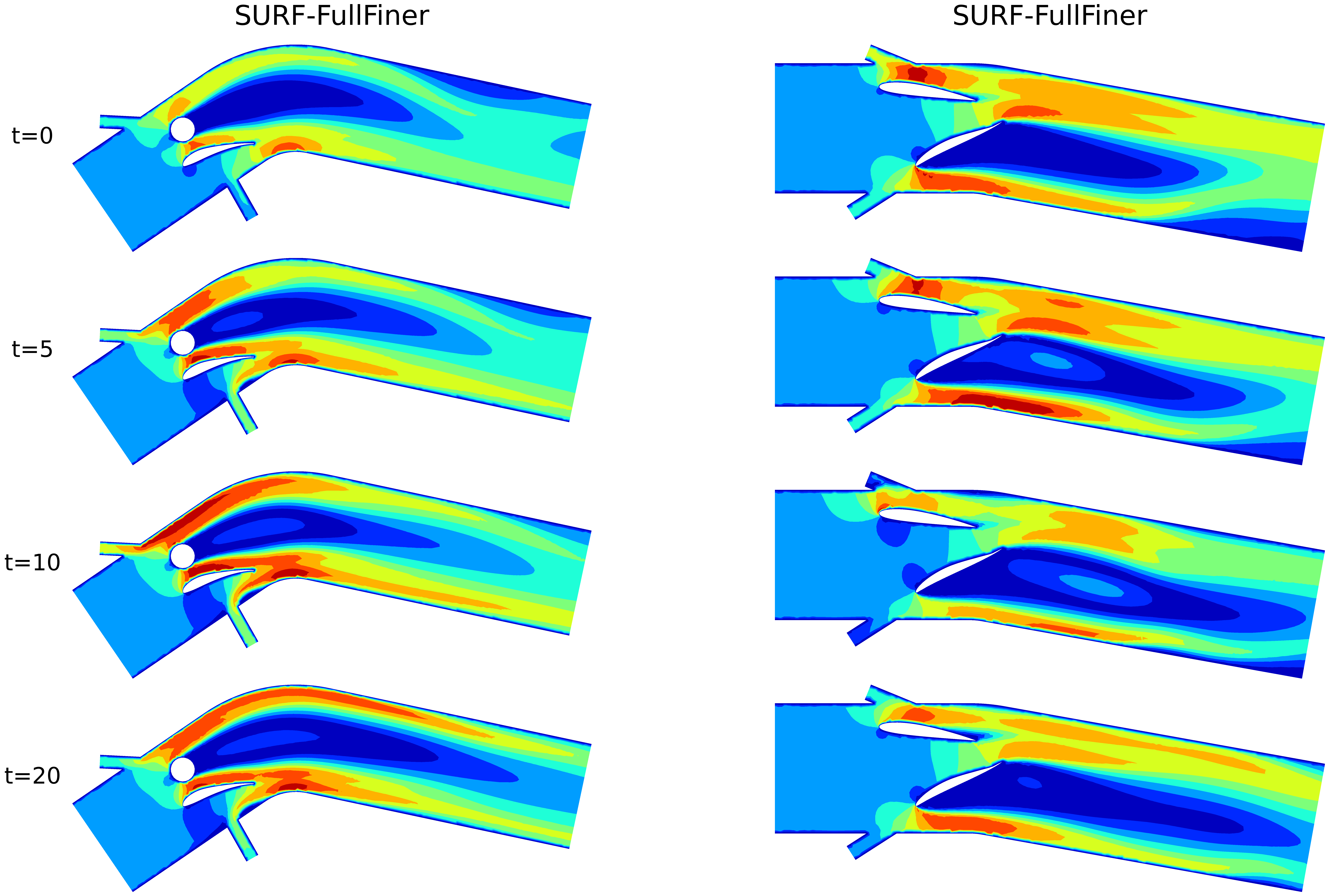}
    \caption{Velocity visualization of two ground-truth datapoints from SURF-Mesh, evaluated at timesteps 0, 5, 10, 20 (top to bottom).}
    \label{fig:FullFinerDatasetExamples}
\end{figure}

Figures \ref{fig:BaseDatasetExamples} to \ref{fig:FullFinerDatasetExamples} depict examples for SURF datapoints. 

\end{document}